\documentclass[11pt]{article}

\usepackage[preprint]{acl}

\usepackage{times}
\usepackage{latexsym}

\usepackage[T1]{fontenc}

\usepackage[utf8]{inputenc}

\usepackage{microtype}

\usepackage{inconsolata}

\usepackage{graphicx}
\usepackage{booktabs}
\usepackage{amssymb}
\usepackage{pifont}
\usepackage{float}
\usepackage{multirow}
\usepackage[table,dvipsnames]{xcolor}
\usepackage{enumitem}
\usepackage{makecell}
\usepackage[most]{tcolorbox}
\usepackage{subcaption}
\usepackage{xspace}
\usepackage{makecell}
\usepackage{array}
\usepackage{colortbl}

\newcommand{\answer}{A\xspace}
\newcommand{\clarifying}{C\xspace}
\newcommand{\enriching}{E\xspace}
\newcommand{\distracting}{D\xspace}
\newcommand{\irrelevant}{I\xspace}
\newcommand{\parametric}{P\xspace}
\xspaceaddexceptions{+}

\newcommand{\hlgreen}[1]{\colorbox{ForestGreen!10}{#1}}
\newcommand{\hlheavygreen}[1]{\colorbox{ForestGreen!30}{#1}}
\newcommand{\hlred}[1]{\colorbox{BrickRed!10}{#1}}
\newcommand{\hlheavyred}[1]{\colorbox{BrickRed!30}{#1}}
\newcommand{\hlwhite}[1]{\colorbox{white!0}{#1}}

\newcommand{\eg}[0]{\textit{e.g.}\xspace}

\newcommand{\ie}[0]{\textit{i.e.}\xspace}
\newcommand{\vs}[0]{\textit{vs.}\xspace}

\newcommand{\llama}[0]{\texttt{Llama-3.3-70B}\xspace}
\newcommand{\gpt}[0]{\texttt{GPT-4o-2024-08-06}\xspace}
\newcommand{\gptshort}[0]{\texttt{GPT-4o}\xspace}
\newcommand{\gemini}[0]{\texttt{Gemini-2.0-Flash}\xspace}
\newcommand{\lme}[0]{LongMemEval-m\xspace}
\newcommand{\lmb}[0]{Long-MT-Bench+\xspace}

\usepackage{etoolbox}

\title{Memory Makes the Difference: Evaluating How Different Memory Roles Shape Conversational Agents}

\author{
  Yuxin Wang\textsuperscript{1}\thanks{This work is done during the author's internship at Microsoft. Correspondence to: \textit{yuxin.wang.gr@dartmouth.edu} and \textit{nickcr@microsoft.com}.} \quad
  Paul Thomas\textsuperscript{2} \quad
  Zhiwei Yu\thanks{Independent researcher.} \quad
  Yuan Gao\textsuperscript{2} \quad
  Saeed Hassanpour\textsuperscript{1} \quad
  \\ \bfseries
  Soroush Vosoughi\textsuperscript{1} \quad
  Robert Sim\textsuperscript{2} \quad
  Nick Craswell\textsuperscript{2}
  \\[0.5em]
  \normalfont\textsuperscript{1}Dartmouth College
  \quad
  \normalfont\textsuperscript{2}Microsoft
}

\begin{document}
\maketitle
\begin{abstract}
Prior research on memory mechanism in RAG-based conversational system has emphasized how memory is stored and retrieved. However, far less is known about how memories with different functional roles influence response quality. Specifically, how they shape an agent's responses under varying conversational contexts and whether they lead to substantively different response behaviors. Existing evaluations in conversational system are also largely reference-based, insufficiently capturing the nuances in responses that may address users' preferences differently. In this work, we probe the impact of different memory types in shaping agents' responses. We present a fine-grained taxonomy of conversational memory, classify retrieved memories into different role types, and design a user-centric evaluation framework that simulates user perspectives. Through comparative experiments on long-term datasets and frontier LLMs, our analysis reveal many differentiated effects of memories: \eg, clarifying memory improves responses' factual accuracy and constraint awareness, making them more correct and personalized; irrelevant memory reduces topic relevance and degrades constraint awareness. Despite the power of frontier LLMs, these findings shed light on how different memory types can be leveraged to produce more personalized responses and inspire further research in this direction. Our code is available at: \url{https://github.com/audreycs/MemoryEffect}.
\end{abstract}

\section{Introduction}
Conversational memory provides needed context for grounded responses and accurate query understanding~\cite{radlinski2017theoretical}, and it now is a core property of contemporary chat systems like ChatGPT~\cite{openai_gpt5} and Google Gemini~\cite{google_gemini}. At high level, conversation history is logged into an external memory bank to reserve past information. During response, chat agents can retrieve relevant pieces to surface useful memory. Such retrieval-augmented generation (RAG) pipeline has served as a standard approach especially in long-range conversation, overcoming the context length limits of Large Language Models (LLMs) to maintain response coherence and accuracy~\cite{fan2024survey}.

\begin{figure}[t]
    \centering
    \includegraphics[width=\columnwidth]{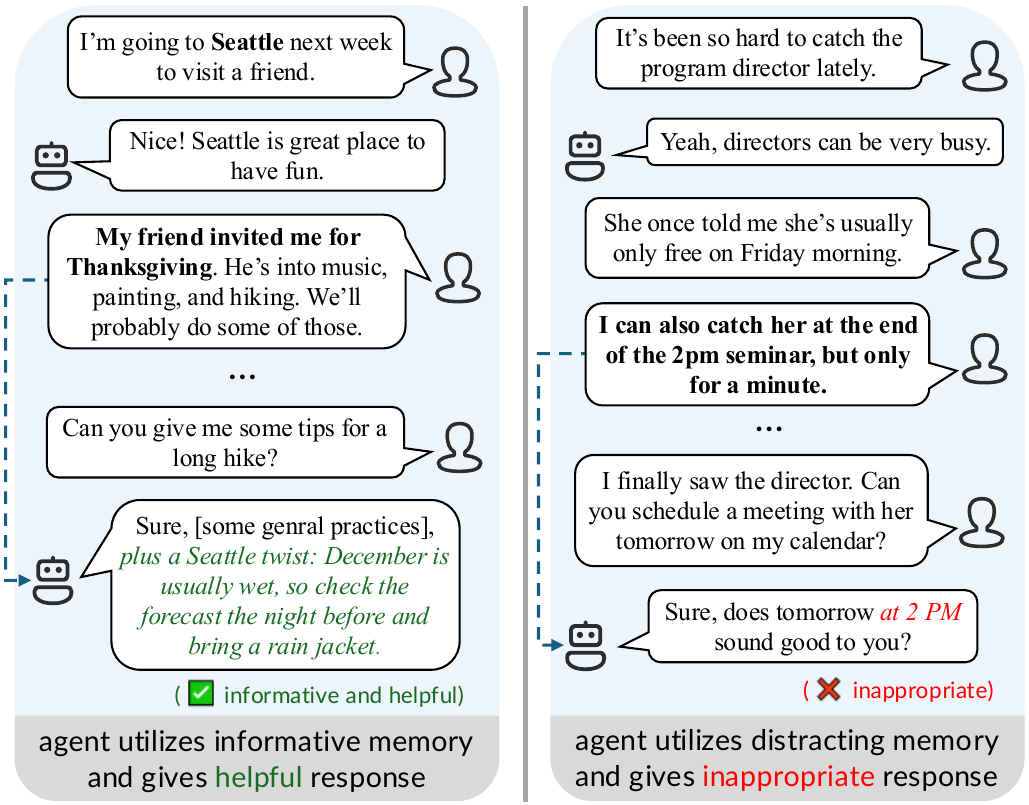}
    \caption{(Left panel) Example of retrieved memory making response more helpful; (Right panel) Example of retrieved memory making response inappropriate.}
    \label{fig:tease}
\end{figure}

Existing work on conversational RAG has largely centered on storage and retrieval strategy, \ie, memory structure, retrieval size, and granularity~\cite{MSC,ye2024boosting,locomo,MemoryBank,wu2025longmemeval,secom,tan-etal-2025-prospect}. They typically retrieve a set of stored pieces, concatenate or compress them into the conversational context, then evaluate the aggregated performance of generation system. 
However, this common practice overlooks the distinct role of properties and impact of individual retrieved pieces. 
Human intuitively recall useful memory to formulate responses, but when it comes to RAG systems, it is uncertain if retrieved memory which are semantically close are actually useful. The effect of retrieval part often behaves like a black box and hard to interpret or control~\cite{benchmarkingllm}. 
As Fig.~\ref{fig:tease} demonstrates, some supplementary memory can enrich the response and make it more helpful; while other memories can be distracting and make the response inappropriate and undesired.
In this work, our central hypothesis is that \textbf{the type of memory matters}: different memory types elicit different agent responses, and these variations can lead to different user perceptions of quality.
Therefore, understanding the qualitative effect of each memory type and whether agents can leverage a bundle of retrieved memories worth systematic investigation. 
Building on this hypothesis, we aim to quantify the impact and find their qualitative property. We introduce the first taxonomy of conversational memory types, determined by their roles when the agent reasons on memory.
Besides, existing work is usually limited in reference-based evaluation. Prior work leans on question answering (QA) datasets provided with unambiguous gold answers~\cite{QReCC,MSC,secom,locomo,chen-etal-2025-compress}. Their reference-based evaluation privileges measuring factual accuracy (\eg, F1, ROUGE) while ignore other qualities of responses like user intent awareness and informativeness, which are critical for dialogic intelligence but difficult to measure using only references.

In this work we present analyses of how different memory types shape LLM agents' responses, paired with a holistic, user-centric evaluation framework covering various aspects. 
Our taxonomy classifies conversational memory pieces into five types --- answer, clarifying, enriching, distracting, and irrelevant, judged with awareness of query and relevant context instead of merely the gold answer in previous analyses~\cite{thepowerofnoise}.
We conducted comparative experiments on two long-term, multi-session conversational datasets over three popular LLMs. Specifically, we compare varied memory settings with (i) answer memory only and (ii) parametric memory only, to study how the presence of individual types and their combinations affects responses quality. To better differentiate response quality, we design a user-centric evaluation proxy that conditions on user intent and prior conversation context, yielding sensitivity to response relevance, informativeness, and other properties. 

Our experiment results reveal many interesting patterns. For examples, we find frontier LLMs are capable of selectively leveraging contextual information with considerable nuance; agents are less likely to ground their responses as retrieval size increases. As for memory types, clarifying memory plays a key role in improving answer accuracy and contextual grounding, and even in isolation it can induce perfect responses. In contrast, irrelevant memory tends to reduce topic relevance and weaken contextual awareness, against the founded positive impact of them on small-scale LLMs like Llama2 and Phi-2 in previous work~\cite{thepowerofnoise}. These also confirm our hypothesis on the impact of memory types. Our taxonomy and user-centric evaluation are general-purpose and can be applied to other conversational corpora and settings. Our main contributions in this work are the following:
\begin{itemize}[topsep=2pt, itemsep=1pt, parsep=0pt]
    \item We hypothesize the impact of memory types on agent's responses, and conducted extensive comparative experiment on frontier LLMs to investigate.
    \item We propose the first fine-grained taxonomy of conversational memory based on functional roles, together with a holistic framework with diverse metrics for generation evaluation.
    \item Our experiments qualitatively reveal the distinct impacts of different memory, with several effects diverging from prior findings on small-scale LLMs.
\end{itemize}

\section{Related Work}
\subsection{Memory-enabled Conversational System}
Conversational systems typically frame memory recall as a store-retrieve-generate pipeline, reflecting flexible memory management and working context~\cite{gao2023neural}. Recent research efforts on LLM-powered conversational systems have largely focused on memory structure~\cite{memgpt,rasmussen2025zep,mem0}, retrieval strategy~\cite{tan2025prospectretrospectreflectivememory,li-etal-2025-hello,hu-etal-2025-hiagent}, and memory update governance~\cite{tan2025prospectretrospectreflectivememory,rezazadeh2025collaborativememorymultiusermemory,wang-etal-2025-unveiling-privacy}. Their retrieval often use sparse or conversational dense retrievers, such as BM25~\cite{bm25} and Contriever~\cite{contriever}. To stress-test these systems in realistic settings, several work target long-term, multi-session dialogue, including MT-Bench-Plus~\cite{memochat}, LOCOMO~\cite{locomo}, and LongMemEval~\cite{wu2025longmemeval}, InfiniteBench~\cite{infinitebench}.

\subsection{Analysis of Retrieval Pieces Type} \label{sec:related_work_type}
Several recent studies have examined how irrelevant or distracting retrieved pieces affect QA performance, primarily focusing on smaller scale LLMs. \citet{thepowerofnoise} give pioneer analysis of how the type and position of retrieved documents influence model's answers. They classified documents into \text{gold}, \text{relevant}, \text{distracting}, and \text{random} based on semantic similarity and the presence of gold answer. They highlighted that random documents can sometimes raise accuracy. \citet{jin2025longcontext} investigate \text{distracting} documents that are relevant to the query but not the gold answer and find consistent harm to QA performance across diverse models. Later, \citet{amiraz-etal-2025-distracting} propose a measure to quantify the distracting effect of distracting documents.
Collectively, these work characterize the impact of irrelevant and distracting evidence in document QA, but give less attention to conversational settings. 
Inspired by their analysis, we adapt and refine their types to conversational scenario and shift the focus from document QA to memory-enabled dialogue, where the responses require more nuanced measurement and multiple forms of ``relevant memory'' (\eg, clarifying, enriching) also need to be considered.

\subsection{Evaluation of Conversational Agent}
Conventional evaluation of conversational system scores responses against references with n-gram or embedding metrics like F1, ROUGE~\cite{rouge}, or BERTScore~\cite{BERTScore}. Newer work on long-term, memory-enabled conversations also tracks retrieval accuracy by measuring if particular pieces are retrieved (\ie, needle in the haystack) via metrics like Recall@$k$~\cite{locomo,wu2025longmemeval}. As the advance of LLM-as-a-judge, LLM-based judges start to be widely adapted for holistic evaluation of chat quality~\cite{mt-bench-101,personalens,mem0}. By setting well-crafted rubrics, LLM judges are flexible to measure diverse dimensions even without references. Mem0~\cite{mem0} prompts LLM judge to compare responses with references, and MT-Bench-101~\cite{mt-bench-101} uses pairwise comparison of two candidate responses for ranking.
PersonaLens~\cite{personalens} equips judge with user profiles and rich interaction history to further measure if responses address user preferences.

\section{Taxonomy of Conversational Memory} \label{sec:memory_taxonomy}
\begin{table}[!t]
    \centering
    \renewcommand{\arraystretch}{1.5}
    \small
    \setlength{\tabcolsep}{1pt}
    \resizebox{\columnwidth}{!}{
    \begin{tabular}{m{3.7cm} m{6.2cm}}
    \Xhline{0.8pt}
    \rowcolor{gray!30}
    \textbf{Memory Type (Abbr.)} & \textbf{Definition} \\
    \Xhline{0.8pt}
    Parametric memory (\parametric)     & Exists in LLMs' parameters through training \\ \hline
    Answer memory (\answer)     & Contains the information of the desired answer for user’s query \\
    \hline
    Clarifying memory (\clarifying)     & Resolves the identity of the asked-for entity or event in the query \\
   \hline
    Enriching memory (\enriching)     &  Makes the final answer more informative or helpful \\
    \hline
    Distracting memory (\distracting)     & Seems to be relevant but is actually misleading \\
    \hline
    Irrelevant memory (\irrelevant)    & Has no meaningful relevance to the user’s query (off-topic) \\
    \Xhline{0.8pt}
    \end{tabular}}
    \caption{Our taxonomy of conversational memory types.}
    \label{tab:memory_taxonomy}
\end{table}

\begin{figure}[t]
    \centering
    \includegraphics[width=\columnwidth]{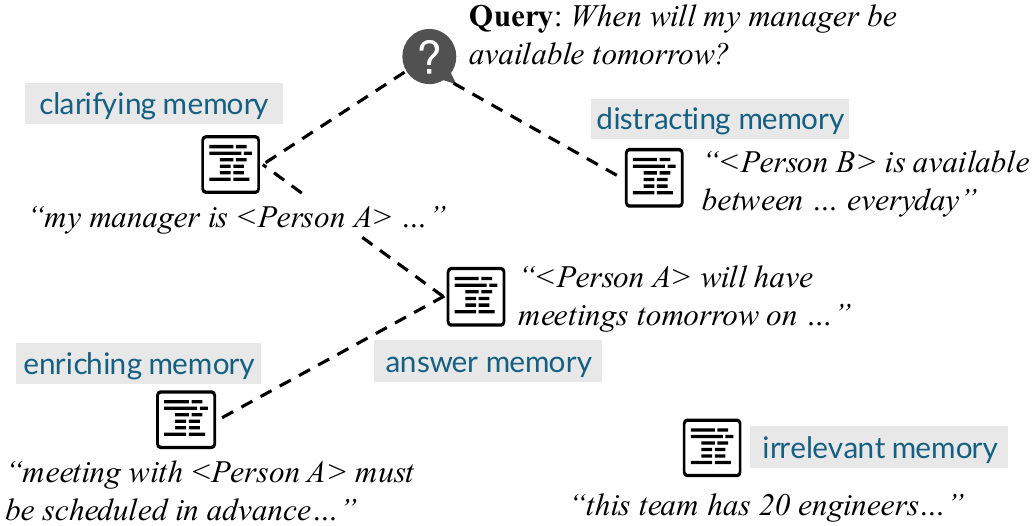}
    \caption{Illustration of relations between memory pieces and user query in agent's reasoning space. Links stand for sematic connection between two.}
    \label{fig:memory_network}
\end{figure}

Retrieved memories form a semantic neighborhood around the query by RAG mechanism, and the agent's reasoning can be viewed as selectively traversing retrieved memory pieces to assemble the information needed for a response, as illustrated in Fig.~\ref{fig:memory_network}.
Based on this intuition and existing document RAG classification~\cite{thepowerofnoise,amiraz-etal-2025-distracting}, we classify conversational memory according to the functional roles it plays during the agent's reasoning process into five fine-grained types: \textit{answer}, \textit{clarifying}, \textit{enriching}, \textit{distracting}, and \textit{irrelevant}. Tab.~\ref{tab:memory_taxonomy} shows their definitions. As shown in Fig.~\ref{fig:memory_network}, to correctly answer the query, the agent needs \text{clarifying} memory to identify the referenced ``manager'', then infer answer from the \text{answer} memory. Ideally, the agent needs  \text{enriching} memory to add more helpful details, and avoids \text{distracting} memory to mitigate false answer. The remaining off-topic memory is considered as \text{irrelevant}, which should be neutral noise. 
Each memory is assigned a single type per query. We acknowledge that, in practice, a memory may plausibly contain pieces serving multiple roles. To enable controlled analysis and clear attribution of effects, we adopt single-label assignment as a deliberate methodological simplification. To resolve overlaps consistently, we apply a labeling precedence: \text{answer} $\rightarrow$ \text{clarifying} $\rightarrow$ \text{enriching} $\rightarrow$ \text{distracting} $\rightarrow$ \text{irrelevant} for disambiguation. 
A practical motivation for the precedence is to encourage sufficient representation of relatively sparse types in realistic retrieval pools. Importantly, it does not encode a hierarchy of importance or quality of them on response generation.
Besides conversational memory, we also include \textit{parametric} memory which is LLM's inherent knowledge learned during its training.

To identify types, prior work~\cite{thepowerofnoise, amiraz-etal-2025-distracting} flags distracting or irrelevant retrieved content by string-matching with the gold answer. While it is efficient and effective for short and unambiguous targets (\eg, a name or location), it falls short for open-ended dialogue responses which are typically more expansive and lexically diverse. Therefore, we instead use a context-aware LLM-as-a-judge to simulate the user and classify the retrieved memory pieces. For each candidate memory, the judge receives the query, the candidate piece, and the top $50$ retrieved memory as the relevant interaction history. This allows the judge to better understand the query and user's intent. Fig.~\ref{fig:prompt_memory_type} presents the details where we adapted the prompt schema from \citet{llms_searcher_preference} which has been shown to be highly effective. We use \texttt{GPT-5-thinking} model~\cite{openai2025chatgpt} as the judge. As memory classification is the cornerstone for following analyses, we also conducted human verification of the classification results, which shows LLM judge achieves human-level reliability (Appendix~\ref{app:human_verification}).

\begin{table*}[t]
    \centering
    \small
    \renewcommand{\arraystretch}{1.5}
    \setlength{\tabcolsep}{4pt}
    \resizebox{\textwidth}{!}{
    \begin{tabular}{l l l c c}
    \Xhline{0.8pt}
    \textbf{Aspect} & \textbf{Metric} & \textbf{Description} & \textbf{Range} & \textbf{Context-aware}\\ \Xhline{0.8pt}
    \multirow{3}{*}{Accuracy} & World Knowledge & Accuracy of the response with respect to general facts and world knowledge & $\{0, 1\}$ & \ding{55} \\ \cline{2-5}
                              & Answer Accuracy & Agreement of the response with the gold answer & $\{0, 1\}$ & \ding{55} \\  \cline{2-5}
                              & Internal Coherence & Logical consistency of the response within itself & $\{0, 1\}$ & \ding{55} \\ 
    \cline{1-5}
    \multirow{2}{*}{Relevance} & Topic Relevance & Whether the response directly addresses the query  & $\{0, 1\}$ & \ding{51} \\ \cline{2-5}
                               & Constraint Awareness & Whether the response satisfies constraints stated or implied in prior interactions & $\{0, 1\}$ & \ding{51} \\  
    \cline{1-5}
    \multirow{2}{*}{Informativeness} & Completeness & Extent to which the response covers relevant information for the query & $\{0, 1, 2\}$ & \ding{51} \\ \cline{2-5}
                                     & Actionability & Whether the response provides concrete details or clear guidance & $\{0, 1\}$ & \ding{55} \\ 
    \cline{1-5}
    & Memory Coverage & Number of retrieved memory pieces whose information is reflected in the response & $\mathbb{Z}^{+}$ & \ding{51} \\ 
    \cline{1-5}
    & ROUGE-1 Recall & Proportion of unigrams in the gold answer that appear in the response & $[0, 1]$ & \ding{55} \\ 
    \Xhline{0.8pt}
    \end{tabular}
    }
    \caption{User-centric evaluation framework}
    \label{tab:eval_framework}
\end{table*}

\section{User-centric Response Evaluation} \label{sec:evaluation}
To better capture the sensitivity of responses to different context, we design a holistic evaluation framework which is user-centric and context-aware (Tab.~\ref{tab:eval_framework}). This framework spans three major aspects (Accuracy, Relevance, and Informativeness) and each aspect contains finer metrics. These aspects are designed by integrating established evaluation framework~\cite{liang2023holistic,maurya-etal-2025-unifying} and adapted to conversational scenario. At a high level, Accuracy captures factual correctness of responses; Relevance checks alignment with user intent and preferences; Informativeness assesses usefulness beyond the minimal answer. Note that conventional reference-based accuracy metric such as ROUGE is subsumed by Answer Accuracy, which tests whether the response contains the desired answer. We additionally include Memory Coverage and ROUGE-1 Recall in the framework. Tab.~\ref{tab:eval_framework} also indicates which metrics require prior conversation history (``context-aware'' column). For instance, World Knowledge Accuracy and Internal Consistency can be judged without context.

For robustness, most metrics except Completeness are designed with a binary scale rather than a fine-grained rating scale. There is work showing binary judgments yield higher accuracy and reliability for LLM judges~\cite{dong-etal-2024-llm,alhawasi2024using}, and finer-grade rating can introduce subjectivity in scale design~\cite{bansal2024peering,llms-are-not-fair}. Thus, we believe binary scale is sufficient for analyses in this work. We use \texttt{GPT-o3} as the LLM judge, and metrics under the same aspect are judged in group.

\section{RAG Pipeline Implementation}
Our implementation of conversational RAG follows the standard RAG pipeline. Some key terms in our context:

\noindent \textbf{Conversation} The complete set of interactions between \underline{one user} and the chat agent. Each user has a separate conversation, which may span multiple sessions over time.

\noindent \textbf{Session} A continuous interaction episode between a user and the agent. A session typically focuses on a single topic.

\noindent \textbf{Utterance and Turn} One utterance is a single text input from either the user or the agent. A turn is a back-and-forth exchange consisting of one user utterance followed by one agent utterance.

\noindent \textbf{Context} For each user query, the context is the relevant prior memory which is reachable to the agent. We refer retrieved memories as context.

\subsection{Two Long-Term Conversational Datasets}
We use two long-term, multi-session conversation datasets for our analyses:  \lme~\cite{wu2025longmemeval} and \lmb~\cite{long-mt-bench-plus}, with statistics in Tab.~\ref{tab:dataset_stats}. \lme contains synthetic conversations paired with human-designed queries. \lmb contains human-written conversations and GPT-generated long-range queries. Both datasets feature extensive conversational memory and target on evaluating memory-related user queries. \lme additionally includes metadata such as session timestamps and evidence memories for each query. 

\begin{table}[t]
    \centering
    \small
    \renewcommand{\arraystretch}{1.3}
    \setlength{\tabcolsep}{3pt}
    \resizebox{\columnwidth}{!}{
    \begin{tabular}{l|rr}
    \Xhline{0.8pt}
    & \textbf{LongMemEval-m}   & \textbf{Long-MT-Bench+} \\
    \Xhline{0.8pt}
    total \# conversation & $500$ & $11$ \\
    \hline
    total \# query    & $500$ & $288$ \\
    \hline
    avg. \# query/conversation & $1$ & $26.2$ \\
    \hline
    avg. \# session/conversation & $501.9$ & $4.9$ \\
    \hline
    avg. \# utterance/conversation & $9.8$ & $26.7$ \\
    \hline
    avg. \# token/conversation & $776,472.4$ & $14,000.9$ \\
    \Xhline{0.8pt}
    \end{tabular}}
    \caption{Statistics of two long-term conversation datasets for experiment. Token counts are measured as space-separated chunks.}
    \label{tab:dataset_stats}
\end{table}

\subsection{Memory Storage and Retrievers}
We use one memory bank for all sessions under the same conversation. When the agent responds to a query, all past sessions are reachable, but different conversations cannot access each other's memory. Following LongMemEval, we implement the memory bank as key-value pairs \{<key>: <value>\}, where the key is used for retrieval indexing and the value returns the original content. We \textbf{store each turn} as one pair; both the key and value are the original turn in our implementation, although the key can be customized (\eg, a summary or a list of keywords). When timestamps are available, we also record them in the values. For retrieval, we experiment with three methods: BM25, Contriever, and OpenAI's \texttt{text-embedding-3-small}~\cite{openaiembedding}. We retrieve top $100$ memory pieces with highest similarity. On average, a memory piece contains $317$ tokens in \lme and $216$ tokens in \lmb, so $100$ pieces already constitute a very large context.

\begin{figure*}[t]
    \vspace{-2mm}
    \centering
    \includegraphics[width=\linewidth]{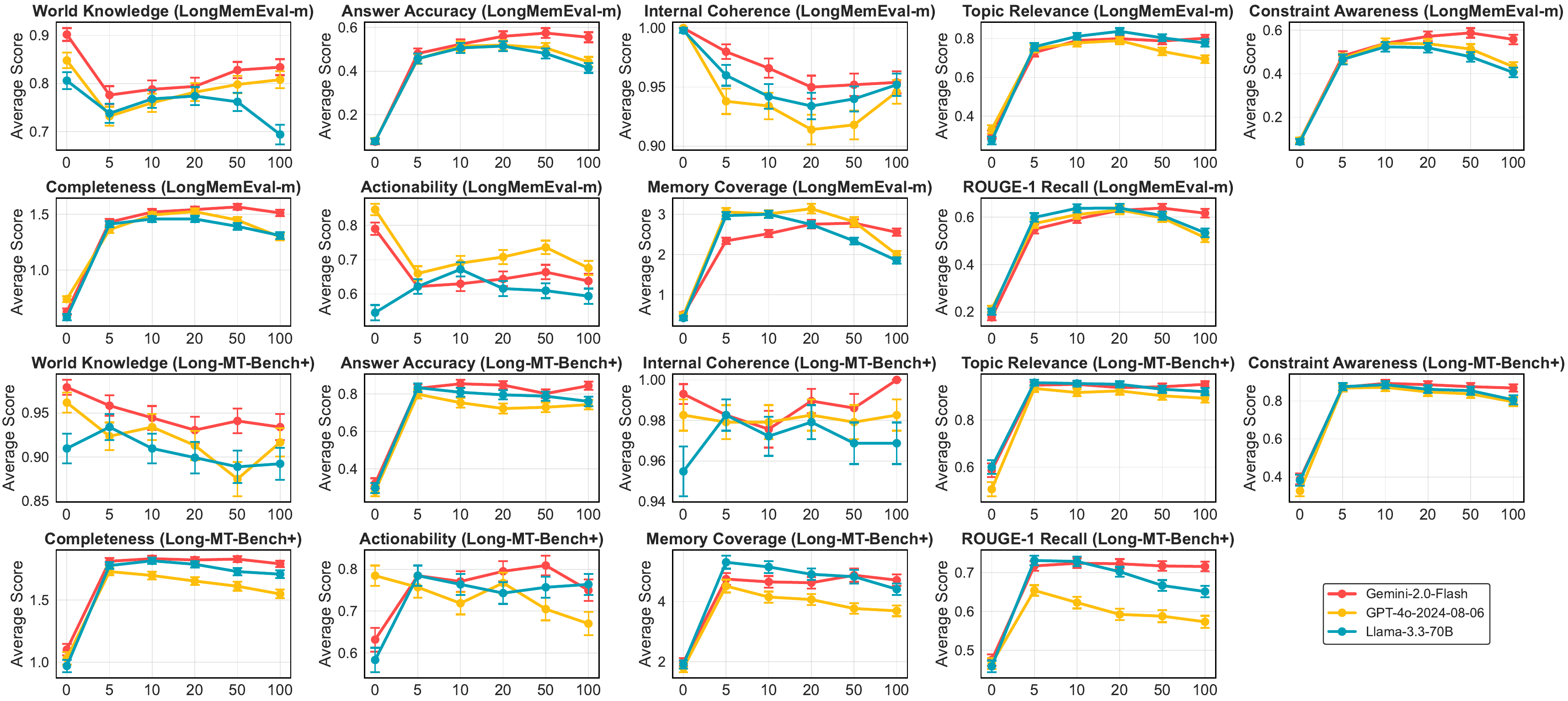}
    \vspace{-5mm}
    \caption{Average metric scores with error bars over context-aware or reference-aware metrics of three retrieval methods on \lme and \lmb datasets. Results are based on \gpt agent. X-axis is the context size used in the prompt (top $k$ memory pieces).}
    \label{fig:general_perf}
\end{figure*}

\begin{figure*}[!ht]
    \centering
    \includegraphics[width=\linewidth]{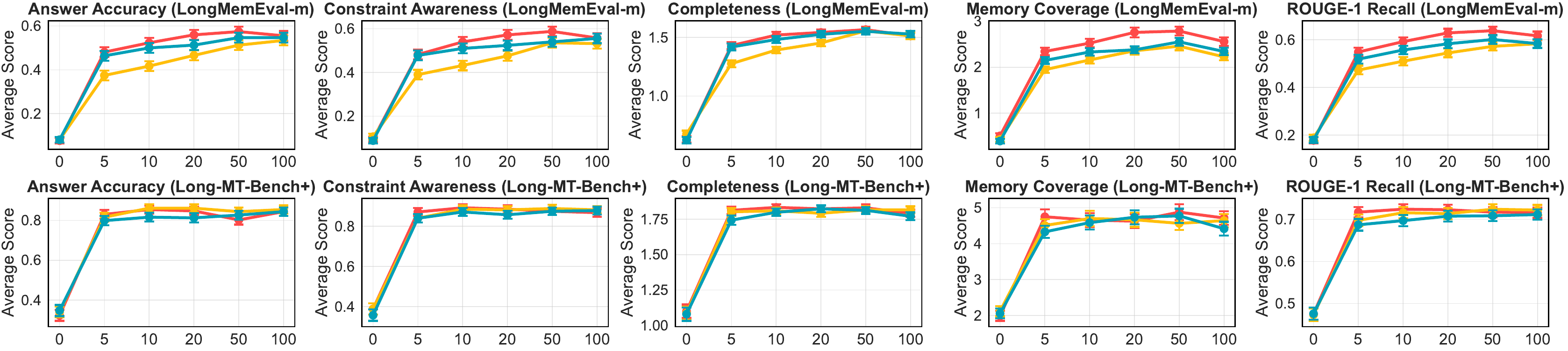}
    \vspace{-5mm}
    \caption{Average scores over context-aware or reference-aware metrics of three retrieval methods on \lme and \lmb datasets. Results are based on \gpt agent. X-axis is the context size used
in the prompt (top $k$ memory pieces).}
    \label{fig:retriever_perf}
    \vspace{-1mm}
\end{figure*}

\subsection{Agent Models and Response Generation}
\looseness=-1
We especially evaluate three large scale and popular LLMs as agents: \gpt~\cite{gpt-4o} (we call \gptshort for short in later parts), \gemini~\cite{gemini2024}, and \llama~\cite{llama3}. We select the top $k$ pieces ($k \in \{0, 5, 10, 20, 50, 100\}$) from the $100$ retrieved pieces to vary context size. When timestamps are available, the $k$ pieces are ordered chronologically. We concatenate them and truncate as needed to fit the model's context window. The full prompt for response generation shown in Fig.~\ref{fig:prompt_response_generation} in Appendix. Note that when $k=0$, the agent relies on its parametric memory to respond. All agents are set with the minimum temperature to ensure near-deterministic output.

\vspace{-5pt}
\subsection{Response Evaluation}
We use \texttt{GPT-o3} with high reasoning effort as the judge. Different rubrics are applied to metrics across different aspects. We carefully designed the rubrics and performed sanity check on a sample of queries to validate the quality of the judgments, iteratively refining the rubrics to align the judge's scores with human evaluations (Appendix~\ref{app:llm_as_judge}). Particularly, for the Topic Relevance metric, refusal responses such as ``\textit{I am unable to answer ...}'' receive a score of $0$. 

\section{General Performance On Large LLMs} \label{sec:general_perf}
First, we analyze general performance on each metric over different settings (agents, context size, retrievers), and we examine agents' memory selectivity by type during response generation.
\subsection{Performance Over Datasets, Agents, Context Size, and Retrievers}
Fig.~\ref{fig:general_perf} shows the average scores over queries with varied the context size (top $k$) on different agents. Fig.~\ref{fig:retriever_perf} shows the average scores for different retrievers. Several notable patterns we find:

\noindent \textbf{Context Size-Performance Trade-off} Performance on metrics such as \textbf{Answer Accuracy} and \textbf{Constraint Awareness} generally rises as the context size $k$ increases, then declines. This suggests that excessive context can distract the agents, echoing prior findings on information overload effects~\cite{locomo,secom,wu2025longmemeval}. Among the three agents, \gptshort shows the strongest robustness to long contexts. 

\noindent \textbf{World Knowledge Accuracy and Constraint Awareness Decline} In Fig.~\ref{fig:general_perf}, when agents respond with only parametric memories ($k=0$), they receive higher World Knowledge Accuracy than with retrieval. Our closer inspection suggests that with added conversational context, agents generate more claims and contextualized statements that judges penalize when unsupported, reducing this score. This pattern indicates that while external evidence can ground responses, it also amplifies hallucination risks, consistent with recent studies~\cite{jin2024long,hu-etal-2025-removal}. Constraint Awareness results also show degradation once context exceeds a certain size, even when Answer Accuracy remains relatively stable (\eg, on \lmb). This suggests the importance of careful triage of historical context, especially in preference-heavy tasks such as recommendation.

\noindent \textbf{Memory Coverage Cap} Results on Memory Coverage and Completeness, which evaluate how much of the retrieved memory is reflected in the response, show a clear upper bound. Coverage peaks when $k \in [5,50]$ and then decline. Top-ranked pieces tend to carry the most useful information, while later ones are often less relevant. As context size grows, the inclusion of irrelevant pieces distracts the agent from leveraging the useful ones, leading to a drop in Memory Coverage.

\noindent \textbf{Retriever Comparison} To compare different retrievers, Fig.~\ref{fig:retriever_perf} shows performances on key metrics that require retrieval or references. Overall, OpenAI's \texttt{text-embedding-3-small} has the edge over other two especially on \lme and smaller $k$, indicating it ranking useful memory higher. BM25 surprisingly remains a competitive baseline despite its simplicity, underscoring that lexical match still finds strong anchors. Their performances converge when retrieving enough memory.

\subsection{Utilization of Different Memory Types}
\begin{figure}[t]
    \centering
    \includegraphics[width=0.95\columnwidth]{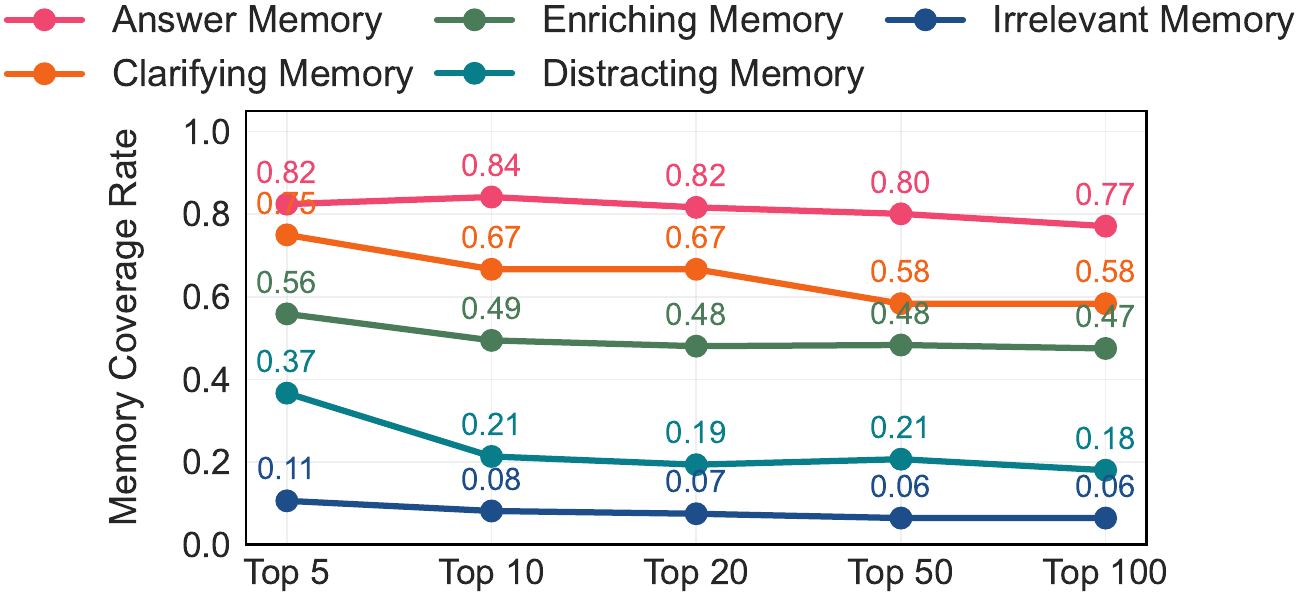}
    \caption{Memory coverage rate of different memory types under different retrieval size $k$. Results are on \gptshort.}
    \label{fig:memory_usage}
\end{figure}

We also measure how often each memory type is covered in the response as the context size $k$ varies. Based on our taxonomy of memory type, ideally the irrelevant memory should be ignored and answer memory should be always included. If this is supported from experiment, it also serves a sanity check for our taxonomy design. To measure this, for a given type, let $m$ be the number of its pieces among the top $k$ retrieved (the \textit{seen} number) and $n$ be the number echoed in the response (the \textit{utilized} number), then the coverage rate is calculated as $n/m$ --- percentage of the seen pieces being used. Fig.~\ref{fig:memory_usage} shows the results on \gptshort. We can see that answer memory is covered most frequently, while irrelevant memory is covered the least, suggesting that the agent is effective in distinguish useful from unuseful memory. Clarifying memory is also highly utilized, suggesting validity of our memory type taxonomy. As $k$ increases, coverage for all types declines slightly. 

\begin{table*}
    \centering
    \small
    \begin{minipage}{\linewidth}
    \renewcommand{\arraystretch}{1.4}
    \setlength{\tabcolsep}{4pt}
    \resizebox{\textwidth}{!}{
    \begin{tabular}{>{\centering\arraybackslash}m{2.6cm}|>{\centering\arraybackslash}m{1.7cm}>{\centering\arraybackslash}m{1.3cm}>{\centering\arraybackslash}m{1.7cm}>{\centering\arraybackslash}m{1.7cm}>{\centering\arraybackslash}m{1.6cm}>{\centering\arraybackslash}m{1.6cm}>{\centering\arraybackslash}m{1.6cm}>{\centering\arraybackslash}m{1.55cm}>{\centering\arraybackslash}m{1.55cm}}
    \Xhline{0.8pt}
    \makecell{\textbf{Comparative}\\\textbf{Experiment} (\vs \answer)} & \textbf{\makecell{World\\Knowledge}} & \textbf{\makecell{Answer\\Accuracy}} & \textbf{\makecell{Internal\\Coherence}} & \textbf{\makecell{Topic\\Relevance}} & \textbf{\makecell{Constraint\\Awareness}} & \textbf{Completeness} & \textbf{Actionability} & \textbf{\makecell{Memory\\Coverage}}
 & \textbf{\makecell{ROUGE-1\\Recall}} \\ \Xhline{0.8pt}
    \answer +\clarifying  & $\hlwhite{0.50} \vrule \hlheavygreen{1.00}$ & $\hlgreen{0.60} \vrule \hlheavygreen{1.00}$ & $\hlwhite{--} \vrule $\hlwhite{--} & $\hlheavygreen{1.00}\vrule\hlheavygreen{1.00}$ & $\hlgreen{0.75}\vrule \hlheavygreen{1.00}$ & $\hlheavygreen{0.89}\vrule\hlheavygreen{1.00}$ & $\hlgreen{0.67}\vrule\hlheavygreen{1.00}$ & $\hlgreen{0.67}\vrule \hlgreen{0.75}$ & $\hlwhite{0.57}\vrule \hlgreen{0.67}$ \\ \hline
    \answer +\enriching & $\hlwhite{0.48} \vrule \hlred{0.28}$ & $\hlwhite{0.53}\vrule \hlgreen{0.65}$ & $\hlred{0.30}\vrule \hlred{0.33}$ & $\hlred{0.33} \vrule \hlheavygreen{0.80}$ & $\hlwhite{0.50} \vrule \hlgreen{0.71}$ & $\hlgreen{0.63} \vrule \hlgreen{0.68}$ & $\hlwhite{0.42} \vrule \hlwhite{0.55}$ & $\hlgreen{0.73}\vrule \hlwhite{0.59}$ & $\hlwhite{0.52} \vrule \hlwhite{0.55}$ \\ \hline
    \answer +\distracting & $\hlred{0.33}\vrule \hlred{0.20}$ & $\hlred{0.42} \vrule \hlwhite{0.50}$ & $\hlred{0.33} \vrule \hlred{0.00}$ & $\hlred{0.33} \vrule \hlwhite{0.50}$ & $\hlred{0.39} \vrule \hlred{0.33}$ & $\hlwhite{0.49} \vrule \hlgreen{0.75}$ & $\hlwhite{0.43} \vrule \hlheavyred{0.00}$ & $\hlgreen{0.65} \vrule \hlgreen{0.73}$ & $\hlred{0.39}\vrule \hlgreen{0.63}$ \\ \hline
    \answer +\irrelevant & $\hlwhite{0.46} \vrule \hlwhite{0.46}$ & $\hlwhite{0.46} \vrule \hlwhite{0.53}$ & $\hlgreen{0.63}\vrule \hlgreen{0.67}$ & $\hlheavyred{0.11} \vrule \hlwhite{0.57}$ & $\hlred{0.38} \vrule \hlred{0.36}$ & $\hlwhite{0.52} \vrule \hlwhite{0.53}$ & $\hlred{0.40}\vrule \hlwhite{0.55}$ & $\hlgreen{0.64}\vrule \hlwhite{0.59}$ & $\hlwhite{0.52}\vrule \hlgreen{0.62}$ \\ \hline
    \clarifying +\enriching +\distracting +\irrelevant & $\hlred{0.34}\vrule \hlred{0.23}$ & $\hlheavyred{0.05}\vrule \hlheavyred{0.10}$ & $\hlgreen{0.67}\vrule \hlwhite{0.50}$ & $\hlheavyred{0.01} \vrule \hlheavyred{0.08}$ & $\hlheavyred{0.03} \vrule \hlheavyred{0.04}$ & $\hlheavyred{0.09} \vrule \hlheavyred{0.20}$ & $\hlred{0.40}\vrule \hlred{0.30}$ & $\hlred{0.37} \vrule \hlred{0.38}$ & $\hlheavyred{0.16} \vrule \hlred{0.22}$ \\ 
 \Xhline{0.8pt}
    \end{tabular}
    }
    \end{minipage}
    
    \vspace{5pt}
    
    \begin{minipage}{\linewidth}
        \renewcommand{\arraystretch}{1.4}
    \setlength{\tabcolsep}{4pt}
    \resizebox{\textwidth}{!}{
    \begin{tabular}{>{\centering\arraybackslash}m{2.6cm}|>{\centering\arraybackslash}m{1.7cm}>{\centering\arraybackslash}m{1.3cm}>{\centering\arraybackslash}m{1.7cm}>{\centering\arraybackslash}m{1.7cm}>{\centering\arraybackslash}m{1.6cm}>{\centering\arraybackslash}m{1.6cm}>{\centering\arraybackslash}m{1.6cm}>{\centering\arraybackslash}m{1.55cm}>{\centering\arraybackslash}m{1.55cm}}
    \Xhline{0.8pt}
    \makecell{\textbf{Comparative}\\\textbf{Experiment} (\vs \parametric)} & \textbf{\makecell{World\\Knowledge}} & \textbf{\makecell{Answer\\Accuracy}} & \textbf{\makecell{Internal\\Coherence}} & \textbf{\makecell{Topic\\Relevance}} & \textbf{\makecell{Constraint\\Awareness}} & \textbf{Completeness} & \textbf{Actionability} & \textbf{\makecell{Memory\\Coverage}}
 & \textbf{\makecell{ROUGE-1\\Recall}} \\ 
    \Xhline{0.8pt}
    \clarifying & $\hlheavyred{0.00}$\vrule $\hlheavyred{0.00}$ & \hspace*{1.5mm}$\hlwhite{--}$\hspace*{1.5mm}\vrule$\hlheavygreen{0.96}$ & $\hlheavyred{0.00}$\vrule $\hlwhite{0.50}$ & $\hlwhite{0.56}$\vrule $\hlheavygreen{1.00}$ & \hspace*{1.5mm}$\hlwhite{--}$\hspace*{1.5mm}\vrule $\hlheavygreen{1.00}$ & $\hlgreen{0.77}$\vrule $\hlheavygreen{0.89}$ & $\hlwhite{0.43}$\vrule $\hlwhite{0.53}$ & $\hlheavygreen{1.00}$\vrule $\hlheavygreen{0.87}$ & $\hlheavygreen{1.00}$\vrule $\hlheavygreen{0.85}$ \\ \hline
    \answer +\clarifying & $\hlheavyred{0.00}$\vrule \hspace*{1.5mm}$\hlwhite{--}$\hspace*{1.5mm} & $\hlheavygreen{0.89}$\vrule $\hlheavygreen{1.00}$ & $\hlwhite{0.50}$\vrule \hspace*{1.5mm}$\hlwhite{--}$\hspace*{1.5mm} & $\hlheavygreen{0.90}$\vrule $\hlheavygreen{1.00}$ & $\hlheavygreen{1.00}$\vrule $\hlheavygreen{1.00}$ & $\hlheavygreen{0.93}$\vrule $\hlheavygreen{0.83}$ & $\hlheavygreen{0.80}$\vrule $\hlheavygreen{1.00}$ & $\hlheavygreen{1.00}$\vrule $\hlheavygreen{1.00}$ & $\hlheavygreen{0.92}$\vrule $\hlheavygreen{1.00}$ \\ \hline
    \enriching & $\hlred{0.26}$\vrule $\hlheavyred{0.04}$ & $\hlheavygreen{0.80}$\vrule $\hlheavygreen{0.91}$ & $\hlheavyred{0.00}$\vrule $\hlheavyred{0.14}$ & $\hlheavygreen{0.82}$\vrule $\hlheavygreen{0.96}$ & $\hlgreen{0.76}$\vrule $\hlheavygreen{0.88}$ & $\hlgreen{0.74}$\vrule $\hlgreen{0.75}$ & $\hlred{0.32}$\vrule $\hlgreen{0.69}$ & $\hlheavygreen{0.85}$\vrule $\hlheavygreen{0.83}$ & $\hlheavygreen{0.82}$\vrule $\hlgreen{0.74}$ \\ \hline
    \distracting & $\hlheavyred{0.20}$\vrule $\hlheavyred{0.08}$ & $\hlheavygreen{0.86}$\vrule $\hlgreen{0.75}$ & \hspace*{1.5mm}$\hlwhite{--}$\hspace*{1.5mm}\vrule $\hlheavyred{0.00}$ & $\hlheavygreen{0.80}$\vrule $\hlheavygreen{0.87}$ & $\hlheavygreen{0.80}$\vrule $\hlgreen{0.77}$ & $\hlgreen{0.70}$\vrule $\hlgreen{0.67}$ & $\hlred{0.27}$\vrule $\hlgreen{0.62}$ & $\hlheavygreen{0.92}$\vrule $\hlheavygreen{0.87}$ & $\hlgreen{0.79}$\vrule $\hlgreen{0.60}$ \\ \hline
    \irrelevant & $\hlred{0.26}$\vrule $\hlheavyred{0.09}$ & $\hlwhite{0.50}$\vrule $\hlwhite{0.51}$ & $\hlheavyred{0.00}$\vrule $\hlred{0.29}$ & $\hlgreen{0.68}$\vrule $\hlgreen{0.64}$ & $\hlwhite{0.43}$\vrule $\hlgreen{0.64}$ & $\hlwhite{0.49}$\vrule $\hlwhite{0.46}$ & $\hlred{0.29}$\vrule $\hlred{0.34}$ & $\hlheavygreen{0.86}$\vrule $\hlgreen{0.72}$ & $\hlgreen{0.75}$\vrule $\hlwhite{0.57}$ \\ \hline
    \answer +\clarifying +\enriching +\distracting +\irrelevant & $\hlred{0.29}$\vrule $\hlheavyred{0.18}$ & $\hlheavygreen{0.96}$\vrule $\hlheavygreen{0.97}$ & $\hlheavyred{0.00}$\vrule $\hlred{0.40}$ & $\hlheavygreen{0.95}$\vrule $\hlheavygreen{0.96}$ & $\hlheavygreen{0.97}$\vrule $\hlheavygreen{0.97}$ & $\hlheavygreen{0.95}$\vrule $\hlheavygreen{0.92}$ & $\hlred{0.28}$\vrule $\hlgreen{0.75}$ & $\hlheavygreen{0.97}$\vrule $\hlheavygreen{0.88}$ & $\hlheavygreen{0.93}$\vrule $\hlheavygreen{0.91}$ \\ 
 \Xhline{0.8pt}
    \end{tabular}
    }
    \end{minipage}
    \caption{Results of improvement rate $\mathrm{IR}$ of comparison experiments on \lme (left number in each cell) and \lmb (right number in each cell) on \gptshort. \textbf{Top table}: experimental setting \vs controlled setting (\answer); \textbf{Bottom table}: experimental setting \vs controlled setting (\parametric). $\mathrm{IR}$ value stands for the chance of experimental setting wins over controlled setting, and $\mathrm{IR}=0.5$ means a tie. ``$-$'' means a lack of data point for the result. We highlight $\mathrm{IR} \in [0.8, 1.0]$ with heavy green, $[0.6, 0.8)$ with light green, $(0.2, 0.4]$ with light red, and $[0, 0.2]$ with heavy red.}
    \label{tab:improvement_rate_answer_gpt}
\end{table*}

\begin{figure*}[t]
    \centering
    \vspace{-8mm}
    \includegraphics[width=\linewidth]{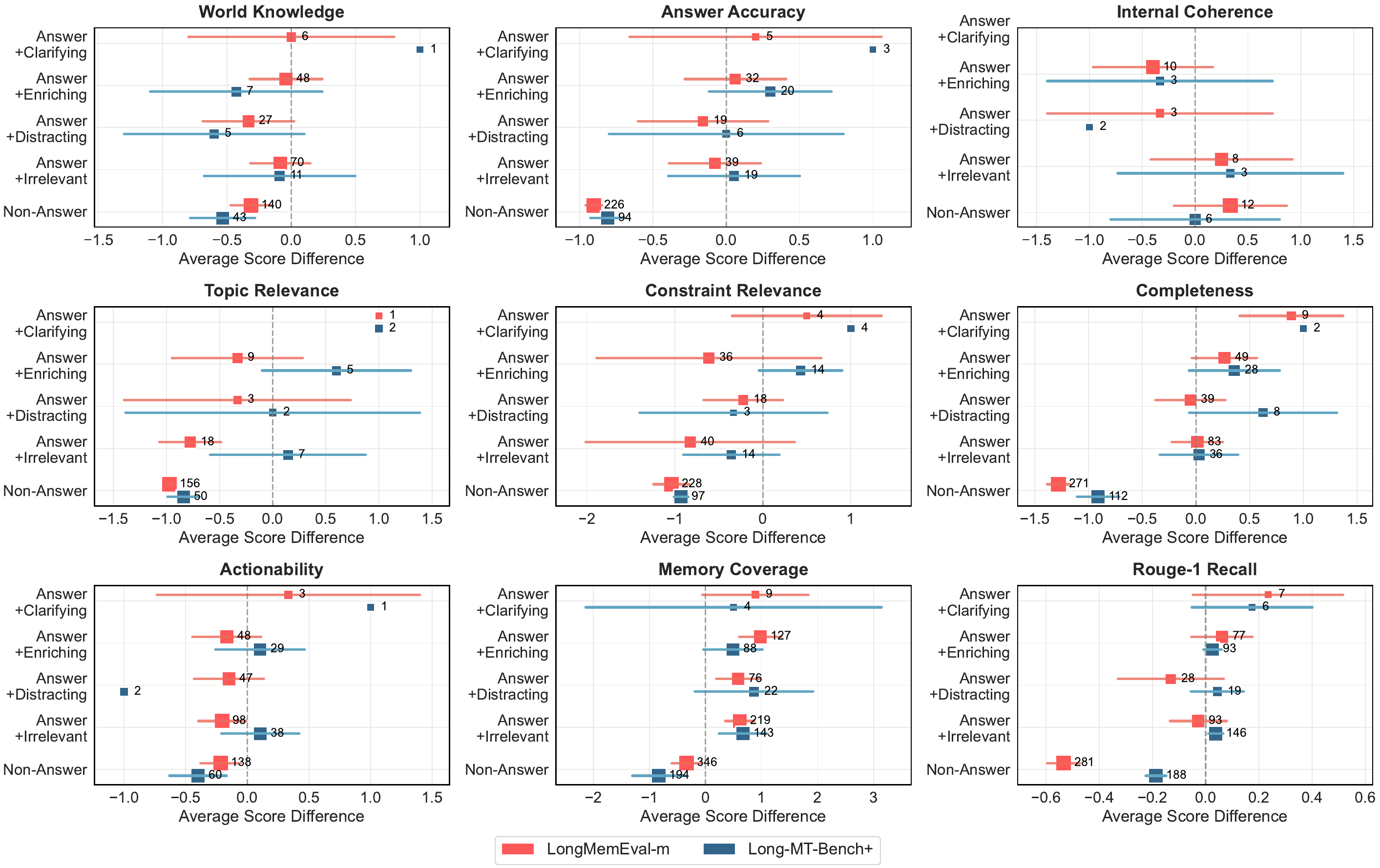}
    \vspace{-5mm}
    \caption{Results of mean per-query score differences with confident intervals when adding a memory setting relative to the answer-only baseline on \gptshort. Squares mark the mean difference; horizontal bars show the confidence interval. Square size is proportional to the number of queries with a non-zero change, and the number shown beside each square equals that count. Absence of square means no qualifying queries for that result.}
    \label{fig:blobbogram_answer_gpt}
    \vspace{-3mm}
\end{figure*}

\section{Impact of Different Memory Types}
Now we take a closer look at how distinct memory types shape response quality.
We conduct controlled experiments aiming to characterize systematic associations between memory-type presence and response behavior under realistic retrieval conditions.
Specifically, we run two sets of comparison experiments: (i) variant settings that add a second memory type on top of an answer-only baseline (\vs \answer); and (ii) variant setting compared to a parametric-only baseline (\vs \parametric). The answer-only baseline reflects a common assumption in RAG: if answer-containing pieces are retrieved in isolation, the model should produce the correct answer easily. In practice, retrieval also contains substantial non-answer memory, so we aim to know how other memories (\clarifying/\enriching/\distracting/\irrelevant) alter responses when present. The parametric-only baseline holds the converse assumption: without memory retrieval, the agent fails to give a good answer to user's queries.
We choose the top $15$ retrieved memory pieces for study (determining their types and varying the memory setting) and test \texttt{text-embedding-3-small} as the retriever. Other experiment setups are the same as above.

\subsection{Comparison with Answer Memory Only}
We compare the \vs \answer baseline to five variants: \answer+\clarifying, \answer+\enriching, \answer+\distracting, \answer+\irrelevant, and a non-answer bundle \clarifying+\enriching+\distracting+\irrelevant. This yields six settings and five pairwise comparisons. As an aside, memory types are judged with query and context, so both the composition and the counts within the top $15$ retrieved pieces can vary. A retrieval may include only one clarifying (\clarifying) piece but more than ten irrelevant (\irrelevant) pieces. Therefore, in comparison settings we retain the original retrieved counts of each memory to reflect the reality. Queries where the retrieved pieces lack answer memory (\answer) are excluded from this analysis as they do not admit a comparable answer baseline.

The idea to quantify each variant setting's impact is checking how its metrics scores change compared to baseline setting.  Formally, let $\mathcal{Q}$ denote the query set for a dataset. For a metric $m$ and a memory setting $i$, let $s^{m}_{q}(i)$ be the score of the agent's response to query $q\in\mathcal{Q}$. We define the per-query score difference between two settings $i$ and $j$ as $\mathrm{diff}^m_q(i,j)$:
{%
\setlength{\abovedisplayskip}{8pt}%
\setlength{\belowdisplayskip}{8pt}%
\begin{equation*}
    \mathrm{diff}^m_q(i,j) = s^m_q(i) - s^m_q(j).
\end{equation*}
}
For instance, $\mathrm{diff}^m_q(\text{\answer}+\text{\clarifying}, \text{\answer}) > 0$ means that adding clarifying memory improves the score on metric $m$ on query $q$ compared to answer-only baseline. We measure how often setting $i$ outperforms setting $j$ on metric $m$ using an indicator \textbf{Improvement Rate} ($\mathrm{IR}$):
{%
\setlength{\abovedisplayskip}{8pt}%
\setlength{\belowdisplayskip}{8pt}%
\begin{equation*}
    \mathrm{IR}^m(i,j) = \frac{\# \{\mathrm{diff}^m_q(i,j) > 0\}}{\# \{\mathrm{diff}^m_q(i,j) \neq 0\}} \in [0,1].
\end{equation*}
}
$\mathrm{IR}$ reflects the directionality of score changes under matched retrieval settings. An $\mathrm{IR}^{m}(i,j)=0.7$ means that, switching from setting $j$ to setting $i$, 70\% of the \textbf{score changes} are towards better. We report $\mathrm{IR}$ for each metric. The top table in Tab.~\ref{tab:improvement_rate_answer_gpt} shows results on \gptshort. Results on other agents are in the Appendix.

Since different memory types distribute vary in top $15$ retrieved pieces, some types (\eg, clarifying) can be scarce while other types (\eg, irrelevant) are relatively more pervasive. This makes the available samples in each variant setting vary and sometimes certain settings only contain few samples for analysis. To further measure the statistical significance of the score changes, we collect the list $L^m(i,j) = \{\mathrm{diff}^m_q(i,j) \neq 0 | q \in \mathcal{Q}\}$ and calculate its mean with $95$\% confidence interval ($\mathrm{ci\_lo}$, $\mathrm{ci\_hi}$). Reporting this helps to contextualize the differences between settings. We visualize the results in Fig.~\ref{fig:blobbogram_answer_gpt} using blobbograms. Below is our key findings:

\noindent \textbf{Significance of Clarifying Memory} The top table in Tab.~\ref{tab:improvement_rate_answer_gpt} shows that responses with clarifying memory (\clarifying) tend to show large gains across both datasets, especially on metrics under Relevance and Informativeness aspects; this is corroborated by the consistent positive statistics in Fig.~\ref{fig:blobbogram_answer_gpt} (the first row). Clarifying memory helps bridge the query to the desired answer and make contextual constraints explicit. This suggests that injecting lightweight disambiguation signals (\eg, entity relation cues or slot clarifications) is a reliably high-leverage practice.

\noindent \textbf{Enriching Memory Improves Informativeness} The presence of enriching memory (\enriching) generally makes responses more informative, as reflected in higher Completeness and Memory Coverage. However, Tab.~\ref{tab:improvement_rate_answer_gpt} also shows a small drop in Internal Coherence, though it is not statistically significant from Fig.~\ref{fig:blobbogram_answer_gpt}. Manual inspection indicates the judge penalizes verbosity and unfocused discussion, \ie, responses become richer but less tightly organized.

\noindent \textbf{Distracting Memory is Harmful} From the top table in Fig.~\ref{tab:improvement_rate_answer_gpt}, incorporating distracting memory (\distracting) substantially degrades performance, particularly on Accuracy and Relevance, indicating that superficially similar yet incorrect content can mislead the model.

\looseness=-1
\noindent \textbf{Necessity of Answer Memory} Relative to the non-answer bundle, answer memory remains essential from the results. Without answer memory, agents are more likely to hedge or decline, lowering Informativeness. However, this pattern suggests that \gptshort is reluctant to fabricate answers even when some relevant context (\clarifying/\enriching/\distracting) is present.

\noindent We also observe that \textbf{irrelevant memory is correlated with reduced constraint awareness} while slightly increasing Memory Coverage and ROUGE-1 Recall. Generally, we do not observe adding irrelevant memory helps. This result extends and challenges existing claim on random/irrelevant documents that they can improve answer-containing Accuracy on small LLMs~\cite{thepowerofnoise}. On one hand, we conjecture that large-scale LLMs behave differently from small LLMs. On another hand, irrelevant memory may divert LLM's attention away from contextual constraints.

\subsection{Comparison with Parametric Memory Only}
In the second comparison group comparing with parametric-only baseline, we evaluate six settings: \clarifying, \answer +\clarifying, \enriching, \distracting, \irrelevant, and all types bundle. The $\mathrm{IR}$ results for \gptshort are in the bottom table of Tab.~\ref{tab:improvement_rate_answer_gpt}, with statistical analysis in Fig.~\ref{fig:blobbogram_parametric_gpt}. Below is our key findings:

\noindent \textbf{Relevant Memory Helps Accuracy and Relevance} The bottom table in Tab.~\ref{tab:improvement_rate_answer_gpt} and Fig.~\ref{fig:blobbogram_parametric_gpt} show that adding semantically relevant memory (answer, clarifying, enriching, or distracting) generally improves Answer Accuracy, Topic Relevance, and Constraint Relevance over parametric-only baseline. This suggests the necessity of past history when agents generating responses. Otherwise, the responses are seriously off-topic and inaccurate.

\noindent \textbf{Clarifying Memory Remains Critical} Among the settings, the \answer+\clarifying and all type bundle settings yield the largest average gains, highlighting the central role of answer-containing evidence and disambiguation information; for \gptshort, these settings also correspond to higher Actionability. Adding clarifying memory (\clarifying) alone also brings significant improvement.

\noindent \textbf{Irrelevant Memory Offers Little Improvement} Irrelevant memory pieces provide no clear improvement over parametric-only, yet they are still frequently echoed in responses from Memory Converge --- a sign that the agent tends to echo irrelevant retrieved text without making the answer better.

\noindent \textbf{RAG Can Reduce Fidelity and Coherence} Across models, we observe adding retrieved memory often lowers World Knowledge Accuracy and Internal Coherence relative to using parametric memory alone. Manual analysis suggests this effect arises because additional context encourages more specific claims, which the LLM judge perceives increases the risk of factual errors, whereas parametric-only responses are shorter, less specific, and safer.

\subsection{Non-significant Results}
We also observe settings where the impact of specific memories is statistically non-significant across agent models and datasets. For example, whether adding distracting memory reduces Internal Coherence relative to the answer-only baseline, and how adding enriching memory affect the Topic Relevance. These effects are small and variable from the results, and stronger evidence would be needed to claim a reliable effect.

\section{Conclusion And Future Directions}
In this work, we examine how memory types affect LLMs in conversations using a novel memory taxonomy, a user-centric evaluation, and controlled comparisons. The experiments over agent models, context size, and retrievers confirm known patterns and provide new findings in long-term conversational RAG scenario. In comparison experiments, we find conversational RAG performance is driven not merely by retrieving relevant context, but by retrieving the right functional types of memory. 

These findings point toward memory-type-aware retrieval and selection as a promising direction for IR. First, retrieval systems can benefit from being memory-role aware, enabling category-aware curation, ranking, and diversification rather than treating all retrieved context uniformly. Second, our results motivate the development of collections that evaluate the different roles of grounding in RAG for memory and other corpora. Finally, it is important to develop and validate taxonomies as analytical tools: the most useful taxonomy is not necessarily the most granular, but the one that best helps us understand retrieval behavior and optimize end-to-end outcomes.
\section*{Limitations}
While our work provides many new findings, it also has several limitations that should be acknowledged.

\noindent \textbf{LLM-as-a-Judge Evaluation} ~ Our experiments heavily reply on LLM judge for response scoring and memory type identification. Though we chose advanced models \texttt{GPT-5-thinking} and \texttt{GPT-o3} as the judges and carefully designed the rubric, LLM judge is known to occasionally produce inconsistent or imperfect assessments. This limitation may introduce  uncertainty into the reported results and our analysis.

\noindent \textbf{Computational Cost} ~ Our experiments involve large-scale experiments with proprietary LLMs such as GPT models and Gemini. Reproducing these results requires substantial API usage, which may be costly and limit full replication by others.

\noindent \textbf{Dataset Size and Noise} ~ The two datasets we used \lme and \lmb contain a few hundred test queries (see their statistics in Table~\ref{tab:dataset_stats}). While labeling errors are expected to be rare, any noise in such a small test size can still introduce minor changes in the reported results. Moreover, some of our results are based on very few data points, which limits their statistical significance and should be interpreted with caution.

\bibliography{custom}

\appendix

\newpage
\section*{Appendix}
\section{Prompt for Response Generation}
The full prompts we use for response generation are shown in Figure~\ref{fig:prompt_response_generation}.

\begin{figure}[h]
  \centering
  \small
  \begin{subfigure}{\columnwidth}
    \centering
    \begin{tcolorbox}[
      width=\columnwidth,
      boxrule=0.7pt,
      left=2mm,
      right=2mm,
      top=1mm,
      bottom=1mm,
      colback=white,  
      colframe=black  
    ]
    \textbf{system}: You are a chat assistant. You previously had a conversation with a user. Now the user is sending you a new message and you need to respond to it. I will provide you with some historical chat pieces with dates between you and the user which you may find helpful to give response. \\
    
    \textbf{user}: 
    Historical Chats: \{\textit{retrieved pieces}\} \\ 
    User's message: \{\textit{query}\}\\ 
    Your response:
    \end{tcolorbox}
  \end{subfigure}

  \vspace{1em} 

  \begin{subfigure}{\columnwidth}
    \centering
    \begin{tcolorbox}[
      width=\columnwidth,
      boxrule=0.7pt,
      left=2mm,
      right=2mm,
      top=1mm,
      bottom=1mm,
      colback=white,  
      colframe=black  
    ]
    \textbf{system}: You are a chat assistant. You previously had a conversation with a user. Now the user is sending you a new message and you need to respond to it. I will provide you with some historical chat pieces between you and the user which you may find helpful to give response. \\
    
    \textbf{user}: 
    Historical Chats: \{\textit{retrieved pieces}\} \\ 
    User's message: \{\textit{query}\}\\ 
    Your response:
    \end{tcolorbox}
  \end{subfigure}

  \vspace{1em}

  \begin{subfigure}{\columnwidth}
    \centering
    \begin{tcolorbox}[
      width=\columnwidth,
      boxrule=0.7pt,
      left=2mm,
      right=2mm,
      top=1mm,
      bottom=1mm,
      colback=white,  
      colframe=black  
    ]
    \textbf{system}: You are a chat assistant. Now the user is sending you a new message and you need to respond to it. \\
    
    \textbf{user}: 
    User's message: \{\textit{query}\} \\ 
    Your response:
    \end{tcolorbox}
  \end{subfigure}
  \caption{Prompt for prompting conversational agent to generate response. (\textbf{top}) prompt for memory with timestamp and memory piece size $k>0$; (\textbf{middle}) prompt for memory without timestamp and $k>0$; (\textbf{bottom}) prompt for $k=0$.}
  \label{fig:prompt_response_generation}
\end{figure}

\section{Human Verification On Memory Type} \label{app:human_verification}

We randomly sampled 5 queries from \lme and 5 from \lmb datasets. For each query, we randomly sampled 3 retrieved memory pieces among the top-15 pieces, whose types are predicted by \texttt{GPT-5-thinking} in our pipeline. In total, we have 30 memory pieces for human evaluation.

We designed a user study questionnaire, mirroring the same instruction and rubric used for the LLM judge, providing example tasks and detailed guidance. Each task is a multiple-choice question to classify the memory into one of the five types.  
We recruited 5 human participants from Prolific\footnote{https://www.prolific.com/}, a third-party data annotation platform, for the study. Our recruitment follows the below criteria:
\begin{itemize}[itemsep=0.2em, topsep=0.2em, parsep=0pt]
    \item Native English speakers residing in English-speaking countries;
    \item Holding bachelor degree or above;
    \item High historical approval rate ($\geq 90\%$);
    \item Age between 18 to 80.
\end{itemize}
We rejected participants whose submissions are with unrealistically short completion times and re-recruit. The compensation is at Prolific's recommended rate.

\begin{table}[t] 
    \centering
    \resizebox{0.9\columnwidth}{!}{
    \begin{tabular}{c|cccccc}
    \Xhline{0.8pt}
     & {GPT-5} & {H1} & {H2} & {H3} & {H4} & {H5} \\
    \Xhline{0.8pt}
    {GPT-5}           & --   & 0.417 & 0.618 & 0.512 & 0.578 & 0.532 \\
    {H1} & 0.417  &  --   & 0.426 & 0.245 & 0.441 & 0.573 \\
    {H2} & 0.618  & 0.426 &  --   & 0.514 & 0.615 & 0.524 \\
    {H3} & 0.512  & 0.245 & 0.514 &  --   & 0.467 & 0.384 \\
    {H4} & 0.578  & 0.441 & 0.615 & 0.467 &  --   & 0.538 \\
    {H5} & 0.532  & 0.573 & 0.524 & 0.384 & 0.538 &  --   \\
    \hline
    \textbf{Average} & \textbf{0.531} & \textbf{0.420} & \textbf{0.540} & \textbf{0.425} & \textbf{0.528} & \textbf{0.510} \\
    \Xhline{0.8pt}
    \end{tabular}
    }
    \caption{Pairwise kappa scores among \texttt{GPT-5-thinking} and human participants (H1-H5). \texttt{GPT-5}'s scores suggest it achieves human-level reliability for memory type classification.}
    \label{tab:human_verification}
\end{table}

Using the 30 annotated memory pieces, we calculate the pairwise Cohen's Kappa $k$ as their inter-annotator agreement, where $k$ is standard unweighted and $k = 0$ corresponds to chance-level agreement. The results are shown in Tab~\ref{tab:human_verification}. 
All pairwise $k$ scores between \texttt{GPT-5-thinking} and human annotators are well above chance, and \texttt{GPT-5-thinking} is comparable to human–human agreement.
\section{Evaluation with LLM-as-a-Judge} \label{app:llm_as_judge}
We carefully designed and iteratively refined the prompts used for LLM judges to ensure reliable and consistent evaluation. The design process followed an iterative ``prompt–test–refine'' loop. Starting from a minimal baseline, we tested responses to identify issues such as instruction drift or unstable reasoning. We then applied human-in-the-loop refinement to adjust phrasing and structure, validating each revision on representative cases until the prompts produce desired outputs on a set of human-validated test samples. We provide judgment cases for memory type classification and response evaluation in Table~\ref{tab:judgment_case_1}, \ref{tab:judgment_case_2}, and ~\ref{tab:judgment_case_3}.

We use \texttt{GPT-5-thinking} as the judge for determining memory types, setting its reasoning effort to \underline{medium}, and max completion tokens to \underline{$8,000$}. Figure~\ref{fig:prompt_memory_type} is the full prompt for it. For the response evaluation, we use \texttt{GPT-o3}, setting its reasoning effort to \underline{high}, and max completion tokens to \underline{$10,000$}. Figure~\ref{fig:prompt_accuracy} is the prompt for evaluating Accuracy; Figure~\ref{fig:prompt_relevance} is for evaluating Relevance; Figure~\ref{fig:prompt_memory_coverage} is for evaluating Memory Coverage; Figure~\ref{fig:prompt_informativeness} is for evaluating Informativeness.

\begin{figure*}[!t]
  \centering
  \small
  \begin{tcolorbox}[
    width=\textwidth,
    boxrule=0.7pt,
    left=2mm,
    right=2mm,
    top=1mm,
    bottom=1mm,
    colback=white,  
    colframe=black
  ]
  \textbf{user}: You are a quality rater evaluating the relevance of a memory piece to a user's query. The memory piece is from the user's conversation with a chat assistant. Your task is to assign an integer score on a scale from 0 to 4 based on the following criteria: \\

  Scoring rubric (check in order 4→3→2→1→0): \\
  4 - Answer: the memory piece clearly answers the query and itself contains the answer or one of the answers for the question. \\
  3 - Clarifying: the memory piece is necessary to resolve the identity of the asked-for entity/event (e.g., coreference, alias, where/which/company disambiguation) so that an answer stated elsewhere can be correctly applied. Without this piece, you cannot derive a correct or confident answer from the remaining information. \\
  2 - Enriching: the memory piece refers to the same specific entity/event (or uniquely narrows to it) and adds useful ancillary detail (e.g., who/where/what/when) that would make the final answer more informative or helpful, even though it does not provide the asked-for attribute. Domain context that is clearly connected to the query also counts as Helpful. \\
  1 - Distracting: the memory piece seems to reveal the answer but is actually misleading. \\
  0 - Irrelevant: the memory piece has no meaningful semantic connection to the user's query domain. \\

  Now I will provide you with a user's query, one memory piece, and other pieces that can be helpful for you to judge. Your task is to assign a relevance score for this memory piece, following the rubric above. \\

  Note: The score should reflect the memory piece itself. The other pieces are provided only to help you understand the memory piece better. \\

  \#\# User's Query: \\
  \{\textit{user query}\} \\

  \#\# Memory Piece: \\
  \{\textit{memory piece}\} \\

  \#\# Other Pieces: \\
  \{\textit{other memory pieces}\} \\

  \#\# Your Score: \\

  \end{tcolorbox}
  \caption{Prompt used by the LLM judge to classify retrieved memory pieces.}
  \label{fig:prompt_memory_type}
\end{figure*}

\begin{figure*}[!ht]
  \centering
  \small
  \begin{tcolorbox}[
    width=\textwidth,
    boxrule=0.7pt,
    left=2mm,
    right=2mm,
    top=1mm,
    bottom=1mm,
    colback=white,  
    colframe=black
  ]
  \textbf{user}: You are an objective and fair evaluator. You are provided with a user's query to a chatbot, the chatbot's response, and the gold response. Your task is to judge the accuracy of the chatbot's response based on three aspects: \\

  1. World knowledge accuracy - How accurate are the factual statements in the response with respect to general facts and world knowledge. A response that does not mention world knowledge is not considered inaccurate. \\
  2. Answer accuracy - How accurate is the response with respect to the answer provided in the gold response. \\
  3. Internal consistency - How logically consistent is the response internally. \\

  You should give \textbf{binary scores} for each aspect where 1 means accurate or consistent and 0 means not, and the explanation for the score. \\

  \#\#\#User's query\#\#\# \\
  \{\textit{question}\} \\

  \#\#\#Chatbot's response\#\#\# \\
  \{\textit{agent response}\} \\

  \#\#\#Gold answer\#\#\# \\
  \{\textit{gold answer}\} \\ 

  You should output only three lines. The first line is the score and explanation for World knowledge accuracy, the second line is the score and explanation for Contextual accuracy, and the third line is the score and explanation for Internal consistency. Each line's format should be `Score: <score>; Explanation: <explanation>'. No index is needed. Now tell me your scores:
  \end{tcolorbox}
  \caption{Prompt for LLM judge to determine the Accuracy of responses.}
  \label{fig:prompt_accuracy}
\end{figure*}

\begin{figure*}[!ht]
  \centering
  \small
  \begin{tcolorbox}[
    width=\textwidth,
    boxrule=0.7pt,
    left=2mm,
    right=2mm,
    top=1mm,
    bottom=1mm,
    colback=white,
    colframe=black
  ]
  \textbf{user}: You are an objective and fair evaluator. You are provided with a user's query to a chatbot, the user's intent behind the query, the chatbot's response to the query, and the prior conversation history. Your task is to judge the relevance of the chatbot's response based on two aspects: \\

  1. Topical relevance - How directly does the response address the query? \\
  2. Constraint relevance - How well does the response meet all implicit or explicit constraints and conditions stated or implied by the user's intent and prior conversation? \\

  You should give \textbf{binary scores} for each aspect where 1 means relevant and 0 means not relevant, and the explanation for the score. \\

  \#\#\#User's query\#\#\# \\
  \{\textit{question}\} \\

  \#\#\#User's intent\#\#\#  (optional) \\
  \{\textit{intent}\} \\

  \#\#\#Chatbot's response\#\#\# \\
  \{\textit{agent response}\} \\

  \#\#\#Prior conversation history\#\#\# \\
  \{\textit{context}\} \\ 

  You should output only two lines. The first line is the score and explanation for Topical relevance, the second line is the score and explanation for Constraint relevance. Each line's format should be `Score: <score>; Explanation: <explanation>'. No index is needed. Now tell me your scores:
  \end{tcolorbox}
  \caption{Prompt for LLM judge to determine the Relevance of responses.}
  \label{fig:prompt_relevance}
\end{figure*}

\begin{figure*}[!ht]
  \centering
  \small
  \begin{tcolorbox}[
    width=\textwidth,
    boxrule=0.7pt,
    left=2mm,
    right=2mm,
    top=1mm,
    bottom=1mm,
    colback=white,  
    colframe=black
  ]
  \textbf{user}: You are a helpful and fair judge. I will provide you with: \\

  1. A user's query to a chatbot; \\
  2. The user's intent behind the query; \\
  3. The chatbot's response; \\
  4. The prior conversation history between the user and the chatbot. \\

  Your task is to identify all pieces of information in the chatbot's response that are recalled from the prior conversation history but were *not* restated in the current user query. \\
  - Count each such recalled item. \\
  - Consider paraphrases or indirect references as recalls. \\
  - Ignore generic pleasantries or facts that never appeared in the history. \\

  \#\#\#Prior conversation history\#\#\# \\
  \{\textit{prior conversation history}\} \\

  \#\#\#User's query\#\#\# \\
  \{\textit{query}\} \\

  \#\#\#User's intent\#\#\# \\
  \{\textit{intent}\} \\

  \#\#\#Chatbot's response\#\#\# \\
  \{\textit{agent response}\} \\ 

  You should output two lines. The first line is the count, the second line is explanation listing the recalled items and in which piece they are recalled from. Now tell me your score and explanation:
  \end{tcolorbox}
  \caption{Prompt for LLM judge to determine the Memory Converge of responses.}
  \label{fig:prompt_memory_coverage}
\end{figure*}

\begin{figure*}[!ht]
  \centering
  \small
  \begin{tcolorbox}[
    width=\textwidth,
    boxrule=0.7pt,
    left=2mm,
    right=2mm,
    top=1mm,
    bottom=1mm,
    colback=white,  
    colframe=black
  ]
  \textbf{user}: You are an objective and fair evaluator. You are provided with a user's query to a chatbot, the user's intent behind the query, the chatbot's response to the query, and the prior conversation history. Your task is to judge the informativeness of the chatbot's response based on two aspects: \\

  1. Completeness - How well does the response cover all sub-questions or necessary information relevant to the query and intent? \\
  2. Actionability - How well does the response provide concrete details or clear guidance? \\

  Particularly, for aspect Completeness, this is a more detailed scale: \\
  Score 0 (Absent): No sign the response noticed the user's motivations, constraints, or intents. Response is completely generic or contradicts the context. \\
  Score 1 (Partial): The response realizes some of the user's intent and motivations and mentions at least one contextual element. But it does not fully address the user's intent. \\
  Score 2 (Comprehensive): The response fully addresses the user's intent and motivations, providing a thorough and contextually relevant answer. \\

  For aspect Actionability, this is a more detailed scale: \\
  Score 0 (Not actionable): The response is vague, generic, or purely descriptive. Lacks concrete items. \\
  Score 1 (Actionable): The response provides a clear sequence of actions or names specific items. \\

  You should give an integer score from 0 to 2 for aspect Completeness and binary score for aspect Actionability. You should also provide explanation for the score. \\

  \#\#\#User's query\#\#\# \\
  \{\textit{question}\} \\

  \#\#\#User's intent\#\#\# \\
  \{\textit{intent}\} \\

  \#\#\#Chatbot's response\#\#\# \\
  \{\textit{agent response}\} \\

  \#\#\#Prior conversation history\#\#\# \\
  \{\textit{context}\} \\ 

  You should output only two lines. The first line is the score and explanation for Completeness, the second line is the score and explanation for Actionability. Each line's format should be `Score: <score>; Explanation: <explanation>'. No index is needed. Now tell me your scores:
  \end{tcolorbox}
  \caption{Prompt for LLM judge to determine the Informativeness of response.}
  \label{fig:prompt_informativeness}
\end{figure*}



\section{Memory Type Classification Details}
We basically rely on prompt in Figure~\ref{fig:prompt_memory_type} to determine the memory type. For the experiment on \lmb, since it contains oracle labels for gold pieces, we directly classify the gold pieces as answer memory.

\section{General Performances on Contriever and BM25}
\begin{figure*}[!ht]
\begin{minipage}{\linewidth}
    \centering
    \includegraphics[width=\linewidth]{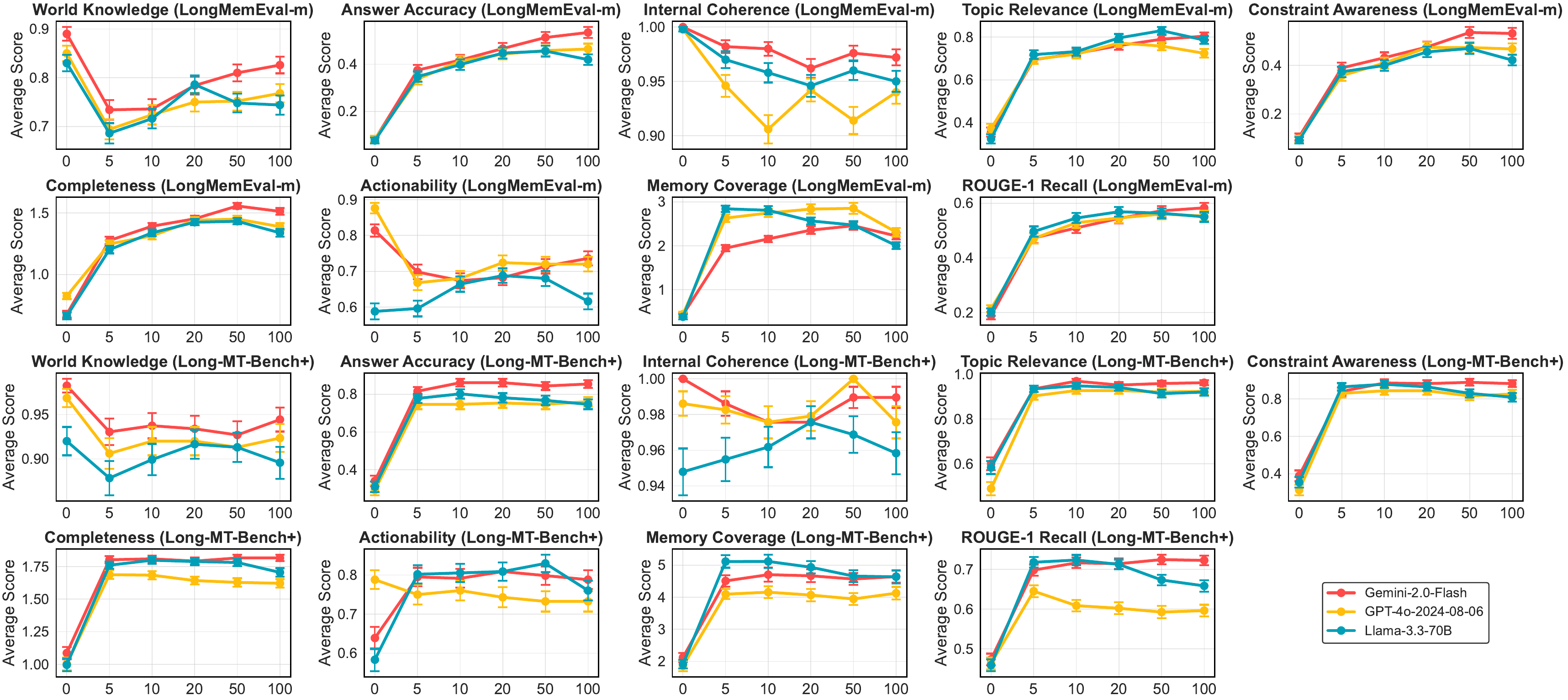}
    \caption{Average scores over different metrics of three agent models on LongMemEval-m and \lmb datasets. Results are based on \texttt{Contriever}. Average scores are calculated over queries.}
    \label{fig:model_comparison_scores_contriever}
\end{minipage}
\begin{minipage}{\linewidth}
    \centering
    \includegraphics[width=\linewidth]{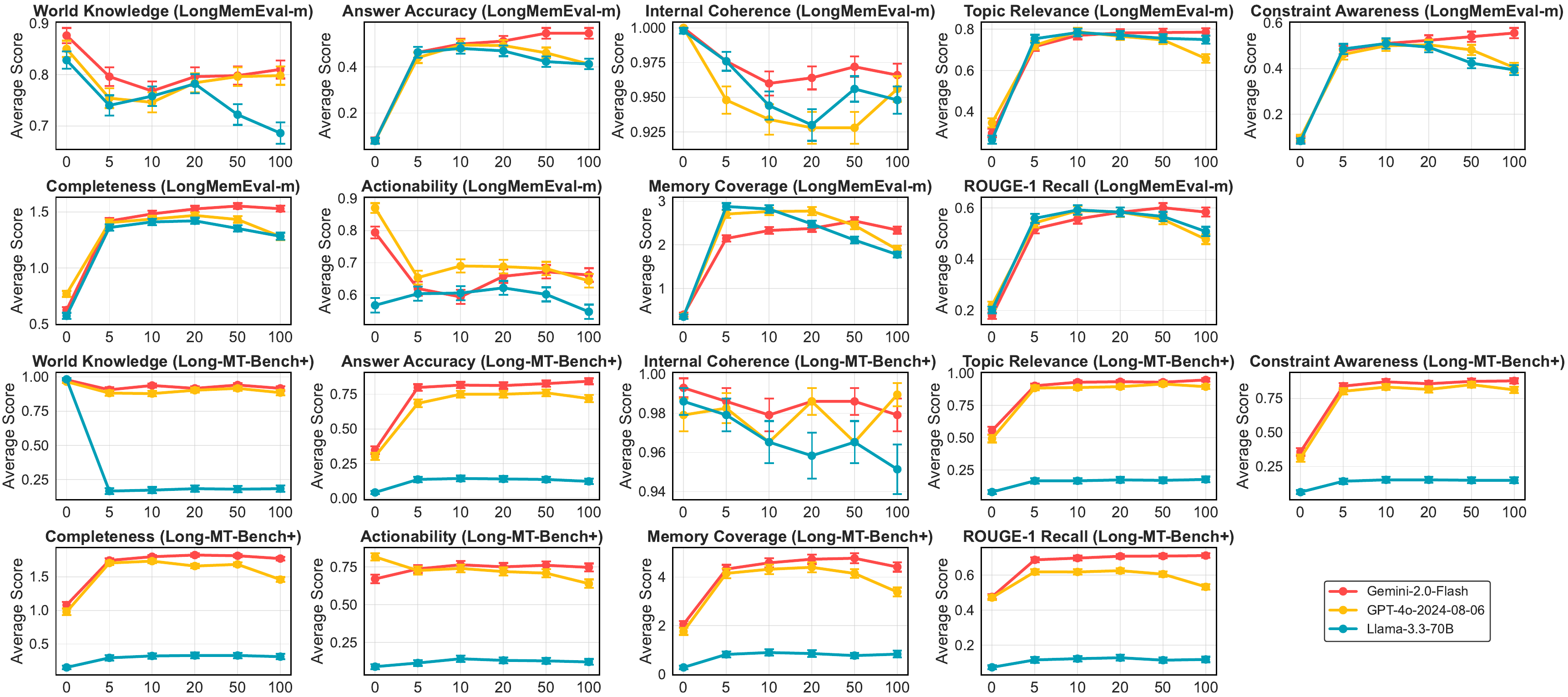}
    \caption{Average scores over different metrics of three agent models on LongMemEval-m and \lmb datasets. Results are based on \texttt{BM25} retriever. Average scores are calculated over queries.}
    \label{fig:model_comparison_scores_bm25}
\end{minipage}
\end{figure*}

Figure~\ref{fig:model_comparison_scores_contriever} and Figure~\ref{fig:model_comparison_scores_bm25} show the general performance over metrics of three LLMs using Contriever and BM25 as retrieval method.

\begin{table*}[!ht]
    \centering
    \small
    \renewcommand{\arraystretch}{1.6}
    \setlength{\tabcolsep}{4pt}
    \resizebox{\textwidth}{!}{
    \begin{tabular}{>{\centering\arraybackslash}m{2.6cm}|>{\centering\arraybackslash}m{1.7cm}>{\centering\arraybackslash}m{1.3cm}>{\centering\arraybackslash}m{1.7cm}>{\centering\arraybackslash}m{1.7cm}>{\centering\arraybackslash}m{1.6cm}>{\centering\arraybackslash}m{1.6cm}>{\centering\arraybackslash}m{1.6cm}>{\centering\arraybackslash}m{1.55cm}>{\centering\arraybackslash}m{1.55cm}}
    \Xhline{0.8pt}
    \makecell{\textbf{Comparative}\\\textbf{Experiment} (\vs \answer)} & \textbf{\makecell{World\\Knowledge}} & \textbf{\makecell{Answer\\Accuracy}} & \textbf{\makecell{Internal\\Coherence}} & \textbf{\makecell{Topic\\Relevance}} & \textbf{\makecell{Constraint\\Awareness}} & \textbf{Completeness} & \textbf{Actionability} & \textbf{\makecell{Memory\\Coverage}}
 & \textbf{\makecell{ROUGE-1\\Recall}} \\ \Xhline{0.8pt}
    \answer +\clarifying  & $\hlgreen{0.60} \vrule ~~\hlwhite{--}~~~$ & $\hlgreen{0.75} \vrule \hlheavyred{0.00}$ & $\hlgreen{0.67} \vrule ~~\hlwhite{--}~~~$ & $\hlheavygreen{1.00} \vrule \hlheavygreen{1.00}$ & $\hlgreen{0.75} \vrule \hlwhite{0.50}$ & $\hlheavygreen{1.00} \vrule \hlred{0.33}$ & $\hlgreen{0.67} \vrule \hlheavyred{0.00}$ & $\hlgreen{0.73} \vrule \hlheavygreen{1.00}$ & $\hlheavygreen{0.83} \vrule \hlheavyred{0.20}$ \\ \hline
\answer +\enriching & $\hlwhite{0.42} \vrule \hlred{0.30}$ & $\hlwhite{0.53} \vrule \hlwhite{0.44}$ & $\hlwhite{0.45} \vrule \hlheavyred{0.00}$ & $\hlred{0.25} \vrule \hlwhite{0.50}$ & $\hlwhite{0.59} \vrule \hlwhite{0.50}$ & $\hlgreen{0.60} \vrule \hlwhite{0.59}$ & $\hlwhite{0.50} \vrule \hlwhite{0.46}$ & $\hlgreen{0.69} \vrule \hlwhite{0.52}$ & $\hlgreen{0.62} \vrule \hlwhite{0.54}$ \\ \hline
\answer +\distracting & $\hlred{0.36} \vrule \hlgreen{0.60}$ & $\hlwhite{0.47} \vrule \hlred{0.40}$ & $\hlgreen{0.71} \vrule \hlheavygreen{1.00}$ & $\hlgreen{0.60} \vrule \hlheavygreen{1.00}$ & $\hlwhite{0.42} \vrule \hlred{0.40}$ & $\hlwhite{0.48} \vrule \hlwhite{0.43}$ & $\hlred{0.40} \vrule \hlheavygreen{0.80}$ & $\hlgreen{0.72} \vrule \hlwhite{0.59}$ & $\hlgreen{0.68} \vrule \hlwhite{0.48}$ \\ \hline
\answer +\irrelevant & $\hlwhite{0.43} \vrule \hlgreen{0.64}$ & $\hlred{0.40} \vrule \hlwhite{0.41}$ & $\hlgreen{0.60} \vrule \hlwhite{0.57}$ & $\hlred{0.30} \vrule \hlred{0.31}$ & $\hlred{0.30} \vrule \hlred{0.36}$ & $\hlwhite{0.52} \vrule \hlred{0.24}$ & $\hlgreen{0.61} \vrule \hlred{0.33}$ & $\hlgreen{0.63} \vrule \hlwhite{0.42}$ & $\hlwhite{0.49} \vrule \hlred{0.40}$ \\ \hline
\clarifying +\enriching +\distracting +\irrelevant & $\hlred{0.27} \vrule \hlheavyred{0.14}$ & $\hlheavyred{0.05} \vrule \hlheavyred{0.04}$ & $\hlwhite{0.55} \vrule \hlwhite{0.55}$ & $\hlheavyred{0.05} \vrule \hlheavyred{0.01}$ & $\hlheavyred{0.06} \vrule \hlheavyred{0.06}$ & $\hlheavyred{0.12} \vrule \hlheavyred{0.12}$ & $\hlred{0.39} \vrule \hlred{0.24}$ & $\hlwhite{0.47} \vrule \hlred{0.29}$ & $\hlheavyred{0.15} \vrule \hlheavyred{0.20}$ \\ \hline
 \Xhline{0.8pt}
    \end{tabular}
    }
    \caption{\gemini's results of $\mathrm{IR}$ of experimental setting (index column) \vs controlled setting (answer memory only) on LongMemEval-m (left number in each cell) and \lmb (right number in each cell).}
    \label{tab:improvement_rate_answer_gemini}
\end{table*}
\begin{table*}[!ht]
    \centering
    \small
    \renewcommand{\arraystretch}{1.6}
    \setlength{\tabcolsep}{4pt}
    \resizebox{\textwidth}{!}{
    \begin{tabular}{>{\centering\arraybackslash}m{2.6cm}|>{\centering\arraybackslash}m{1.7cm}>{\centering\arraybackslash}m{1.3cm}>{\centering\arraybackslash}m{1.7cm}>{\centering\arraybackslash}m{1.7cm}>{\centering\arraybackslash}m{1.6cm}>{\centering\arraybackslash}m{1.6cm}>{\centering\arraybackslash}m{1.6cm}>{\centering\arraybackslash}m{1.55cm}>{\centering\arraybackslash}m{1.55cm}}
    \Xhline{0.8pt}
    \makecell{\textbf{Comparative}\\\textbf{Experiment} (\vs \answer)} & \textbf{\makecell{World\\Knowledge}} & \textbf{\makecell{Answer\\Accuracy}} & \textbf{\makecell{Internal\\Coherence}} & \textbf{\makecell{Topic\\Relevance}} & \textbf{\makecell{Constraint\\Awareness}} & \textbf{Completeness} & \textbf{Actionability} & \textbf{\makecell{Memory\\Coverage}}
 & \textbf{\makecell{ROUGE-1\\Recall}} \\ \Xhline{0.8pt}
    \answer +\clarifying  & $\hlred{0.25} \vrule \hlheavyred{0.00}$ & $\hlgreen{0.60} \vrule \hlheavygreen{1.00}$ & $\hlheavygreen{1.00} \vrule \hlheavyred{0.00}$ & $~~~\hlwhite{--}~~~ \vrule \hlheavygreen{1.00}$ & $\hlwhite{0.50} \vrule \hlheavygreen{1.00}$ & $\hlheavygreen{0.80} \vrule \hlgreen{0.67}$ & $\hlred{0.40} \vrule \hlheavygreen{1.00}$ & $\hlheavygreen{0.90} \vrule \hlgreen{0.67}$ & $\hlgreen{0.78} \vrule \hlwhite{0.50}$ \\ \hline
    \answer +\enriching & $\hlwhite{0.53} \vrule \hlred{0.29}$ & $\hlwhite{0.47} \vrule \hlgreen{0.73}$ & $\hlred{0.38} \vrule \hlheavyred{0.00}$ & $\hlheavygreen{0.83} \vrule \hlgreen{0.67}$ & $\hlwhite{0.48} \vrule \hlwhite{0.53}$ & $\hlgreen{0.60} \vrule \hlgreen{0.65}$ & $\hlwhite{0.54} \vrule \hlwhite{0.59}$ & $\hlgreen{0.71} \vrule \hlwhite{0.59}$ & $\hlgreen{0.61} \vrule \hlgreen{0.61}$ \\ \hline
    \answer +\distracting & $\hlred{0.36} \vrule \hlwhite{0.50}$ & $\hlred{0.39} \vrule \hlwhite{0.50}$ & $\hlgreen{0.64} \vrule \hlwhite{0.50}$ & $\hlheavyred{0.00} \vrule \hlwhite{0.50}$ & $\hlwhite{0.52} \vrule \hlheavygreen{0.80}$ & $\hlwhite{0.43} \vrule \hlwhite{0.57}$ & $\hlwhite{0.48} \vrule \hlgreen{0.75}$ & $\hlgreen{0.60} \vrule \hlgreen{0.62}$ & $\hlwhite{0.54} \vrule \hlgreen{0.64}$ \\ \hline
    \answer +\irrelevant & $\hlwhite{0.46} \vrule \hlwhite{0.50}$ & $\hlred{0.36} \vrule \hlgreen{0.61}$ & $\hlwhite{0.56} \vrule \hlwhite{0.56}$ & $\hlheavyred{0.19} \vrule \hlheavyred{0.00}$ & $\hlred{0.30} \vrule \hlwhite{0.48}$ & $\hlwhite{0.46} \vrule \hlred{0.38}$ & $\hlwhite{0.44} \vrule \hlgreen{0.63}$ & $\hlwhite{0.50} \vrule \hlwhite{0.54}$ & $\hlwhite{0.41} \vrule \hlwhite{0.50}$ \\ \hline
    \clarifying +\enriching +\distracting +\irrelevant & $\hlred{0.24} \vrule \hlheavyred{0.08}$ & $\hlheavyred{0.04} \vrule \hlheavyred{0.06}$ & $\hlwhite{0.44} \vrule \hlwhite{0.57}$ & $\hlheavyred{0.02} \vrule \hlheavyred{0.00}$ & $\hlheavyred{0.02} \vrule \hlheavyred{0.08}$ & $\hlheavyred{0.06} \vrule \hlheavyred{0.13}$ & $\hlheavyred{0.19} \vrule \hlred{0.26}$ & $\hlwhite{0.48} \vrule \hlred{0.33}$ & $\hlheavyred{0.16} \vrule \hlheavyred{0.19}$ \\ \hline
 \Xhline{0.8pt}
    \end{tabular}
    }
    \caption{\llama's results of $\mathrm{IR}$ of experimental setting (index column) \vs controlled setting (answer memory only) on LongMemEval-m (left number in each cell) and \lmb (right number in each cell).}
    \label{tab:improvement_rate_answer_llama}
\end{table*}
\begin{table*}[!ht]
    \centering
    \small
    \renewcommand{\arraystretch}{1.6}
    \setlength{\tabcolsep}{4pt}
    \resizebox{\textwidth}{!}{
    \begin{tabular}{>{\centering\arraybackslash}m{2.6cm}|>{\centering\arraybackslash}m{1.7cm}>{\centering\arraybackslash}m{1.3cm}>{\centering\arraybackslash}m{1.7cm}>{\centering\arraybackslash}m{1.7cm}>{\centering\arraybackslash}m{1.6cm}>{\centering\arraybackslash}m{1.6cm}>{\centering\arraybackslash}m{1.6cm}>{\centering\arraybackslash}m{1.55cm}>{\centering\arraybackslash}m{1.55cm}}
    \Xhline{0.8pt}
    \makecell{\textbf{Comparative}\\\textbf{Experiment} (\vs \parametric)} & \textbf{\makecell{World\\Knowledge}} & \textbf{\makecell{Answer\\Accuracy}} & \textbf{\makecell{Internal\\Coherence}} & \textbf{\makecell{Topic\\Relevance}} & \textbf{\makecell{Constraint\\Awareness}} & \textbf{Completeness} & \textbf{Actionability} & \textbf{\makecell{Memory\\Coverage}}
 & \textbf{\makecell{ROUGE-1\\Recall}} \\ 
    \Xhline{0.8pt}
    \clarifying & $\hlheavyred{0.17} \vrule \hlheavyred{0.12}$ & $\hlheavyred{0.00} \vrule \hlheavygreen{0.91}$ & $\hlheavyred{0.00} \vrule \hlheavyred{0.00}$ & $\hlheavygreen{0.88} \vrule \hlheavygreen{0.92}$ & $~~~\hlwhite{--}~~~ \vrule \hlheavygreen{0.88}$ & $\hlwhite{0.42} \vrule \hlheavygreen{0.81}$ & $\hlred{0.25} \vrule \hlred{0.22}$ & $\hlheavygreen{0.87} \vrule \hlheavygreen{0.87}$ & $\hlwhite{0.58} \vrule \hlgreen{0.76}$ \\ \hline
    \answer +\clarifying & $\hlwhite{0.57} \vrule \hlheavyred{0.00}$ & $\hlheavygreen{0.89} \vrule \hlgreen{0.75}$ & $\hlheavyred{0.00} \vrule ~~~\hlwhite{--}~~~$ & $\hlheavygreen{1.00} \vrule \hlheavygreen{1.00}$ & $\hlheavygreen{1.00} \vrule \hlheavygreen{0.80}$ & $\hlheavygreen{0.93} \vrule \hlheavygreen{0.83}$ & $\hlred{0.40} \vrule \hlheavyred{0.00}$ & $\hlheavygreen{0.88} \vrule \hlheavygreen{0.86}$ & $\hlheavygreen{0.82} \vrule \hlheavygreen{1.00}$ \\ \hline
    \enriching & $\hlheavyred{0.20} \vrule \hlheavyred{0.08}$ & $\hlgreen{0.76} \vrule \hlheavygreen{0.82}$ & $\hlheavyred{0.00} \vrule \hlred{0.40}$ & $\hlheavygreen{0.84} \vrule \hlheavygreen{0.87}$ & $\hlgreen{0.72} \vrule \hlheavygreen{0.88}$ & $\hlgreen{0.62} \vrule \hlgreen{0.75}$ & $\hlheavyred{0.20} \vrule \hlred{0.37}$ & $\hlheavygreen{0.87} \vrule \hlheavygreen{0.84}$ & $\hlgreen{0.63} \vrule \hlgreen{0.69}$ \\ \hline
    \distracting & $\hlred{0.23} \vrule \hlheavyred{0.00}$ & $\hlgreen{0.69} \vrule \hlgreen{0.62}$ & $\hlheavyred{0.00} \vrule \hlheavyred{0.17}$ & $\hlgreen{0.76} \vrule \hlgreen{0.73}$ & $\hlgreen{0.68} \vrule \hlgreen{0.62}$ & $\hlwhite{0.48} \vrule \hlgreen{0.63}$ & $\hlheavyred{0.16} \vrule \hlred{0.35}$ & $\hlheavygreen{0.89} \vrule \hlheavygreen{0.81}$ & $\hlwhite{0.59} \vrule \hlwhite{0.45}$ \\ \hline
    \irrelevant & $\hlred{0.23} \vrule \hlheavyred{0.06}$ & $\hlwhite{0.50} \vrule \hlred{0.39}$ & $\hlheavyred{0.00} \vrule \hlwhite{0.50}$ & $\hlgreen{0.68} \vrule \hlred{0.35}$ & $\hlwhite{0.44} \vrule \hlwhite{0.43}$ & $\hlwhite{0.41} \vrule \hlred{0.22}$ & $\hlheavyred{0.12} \vrule \hlheavyred{0.18}$ & $\hlheavygreen{0.88} \vrule \hlwhite{0.58}$ & $\hlgreen{0.60} \vrule \hlred{0.29}$ \\ \hline
    \answer +\clarifying +\enriching +\distracting +\irrelevant & $\hlred{0.38} \vrule \hlred{0.28}$ & $\hlheavygreen{0.95} \vrule \hlheavygreen{0.92}$ & $\hlheavyred{0.00} \vrule \hlwhite{0.50}$ & $\hlheavygreen{0.93} \vrule \hlheavygreen{0.96}$ & $\hlheavygreen{0.95} \vrule \hlheavygreen{0.96}$ & $\hlheavygreen{0.92} \vrule \hlgreen{0.78}$ & $\hlred{0.29} \vrule \hlwhite{0.47}$ & $\hlheavygreen{0.95} \vrule \hlheavygreen{0.81}$ & $\hlheavygreen{0.89} \vrule \hlgreen{0.67}$ \\ \hline
     \Xhline{0.8pt}
    \end{tabular}
    }
    \caption{\gemini's results of $\mathrm{IR}$ of experimental setting (index column) \vs controlled setting (parametric memory only) on \lme (left number in each cell) and \lmb (right number in each cell).}
    \label{tab:improvement_rate_parametric_gemini}
\end{table*}
\begin{table*}[!ht]
    \centering
    \small
    \renewcommand{\arraystretch}{1.6}
    \setlength{\tabcolsep}{4pt}
    \resizebox{\textwidth}{!}{
    \begin{tabular}{>{\centering\arraybackslash}m{2.6cm}|>{\centering\arraybackslash}m{1.7cm}>{\centering\arraybackslash}m{1.3cm}>{\centering\arraybackslash}m{1.7cm}>{\centering\arraybackslash}m{1.7cm}>{\centering\arraybackslash}m{1.6cm}>{\centering\arraybackslash}m{1.6cm}>{\centering\arraybackslash}m{1.6cm}>{\centering\arraybackslash}m{1.55cm}>{\centering\arraybackslash}m{1.55cm}}
    \Xhline{0.8pt}
    \makecell{\textbf{Comparative}\\\textbf{Experiment} (\vs \parametric)} & \textbf{\makecell{World\\Knowledge}} & \textbf{\makecell{Answer\\Accuracy}} & \textbf{\makecell{Internal\\Coherence}} & \textbf{\makecell{Topic\\Relevance}} & \textbf{\makecell{Constraint\\Awareness}} & \textbf{Completeness} & \textbf{Actionability} & \textbf{\makecell{Memory\\Coverage}}
 & \textbf{\makecell{ROUGE-1\\Recall}} \\ 
    \Xhline{0.8pt}
    \clarifying & $\hlred{0.25} \vrule \hlheavyred{0.00}$ & $~~~\hlwhite{--}~~~ \vrule \hlheavygreen{0.95}$ & $\hlheavyred{0.00} \vrule \hlheavyred{0.00}$ & $\hlheavygreen{0.89} \vrule \hlheavygreen{0.95}$ & $\hlheavygreen{1.00} \vrule \hlheavygreen{0.96}$ & $\hlheavygreen{0.91} \vrule \hlheavygreen{0.93}$ & $\hlgreen{0.60} \vrule \hlwhite{0.50}$ & $\hlheavygreen{0.95} \vrule \hlheavygreen{0.93}$ & $\hlheavygreen{1.00} \vrule \hlheavygreen{0.90}$ \\ \hline
    \answer +\clarifying & $\hlwhite{0.50} \vrule \hlheavyred{0.00}$ & $\hlheavygreen{0.88} \vrule \hlheavygreen{1.00}$ & $~~~\hlwhite{--}~~~ \vrule \hlheavyred{0.00}$ & $\hlheavygreen{0.92} \vrule \hlheavygreen{1.00}$ & $\hlheavygreen{1.00} \vrule \hlheavygreen{1.00}$ & $\hlheavygreen{1.00} \vrule \hlheavygreen{1.00}$ & $\hlheavygreen{0.89} \vrule \hlheavygreen{1.00}$ & $\hlheavygreen{0.94} \vrule \hlheavygreen{1.00}$ & $\hlheavygreen{1.00} \vrule \hlheavygreen{1.00}$ \\ \hline
    \enriching & $\hlred{0.22} \vrule \hlred{0.25}$ & $\hlheavygreen{0.82} \vrule \hlheavygreen{0.83}$ & $\hlheavyred{0.00} \vrule \hlgreen{0.70}$ & $\hlheavygreen{0.90} \vrule \hlheavygreen{0.88}$ & $\hlgreen{0.79} \vrule \hlheavygreen{0.81}$ & $\hlgreen{0.79} \vrule \hlgreen{0.76}$ & $\hlwhite{0.44} \vrule \hlgreen{0.61}$ & $\hlheavygreen{0.96} \vrule \hlheavygreen{0.87}$ & $\hlgreen{0.79} \vrule \hlheavygreen{0.83}$ \\ \hline
    \distracting & $\hlheavyred{0.08} \vrule \hlheavyred{0.08}$ & $\hlwhite{0.53} \vrule \hlgreen{0.67}$ & $\hlheavyred{0.00} \vrule \hlwhite{0.43}$ & $\hlheavygreen{0.84} \vrule \hlgreen{0.75}$ & $\hlwhite{0.57} \vrule \hlwhite{0.57}$ & $\hlgreen{0.62} \vrule \hlgreen{0.77}$ & $\hlred{0.26} \vrule \hlwhite{0.50}$ & $\hlheavygreen{0.96} \vrule \hlgreen{0.76}$ & $\hlgreen{0.70} \vrule \hlgreen{0.73}$ \\ \hline
    \irrelevant & $\hlred{0.25} \vrule \hlheavyred{0.13}$ & $\hlred{0.33} \vrule \hlwhite{0.44}$ & $\hlheavyred{0.06} \vrule \hlgreen{0.65}$ & $\hlgreen{0.74} \vrule \hlwhite{0.47}$ & $\hlred{0.38} \vrule \hlred{0.37}$ & $\hlwhite{0.50} \vrule \hlred{0.39}$ & $\hlred{0.22} \vrule \hlred{0.29}$ & $\hlheavygreen{0.93} \vrule \hlgreen{0.75}$ & $\hlgreen{0.71} \vrule \hlwhite{0.56}$ \\ \hline
    \answer +\clarifying +\enriching +\distracting +\irrelevant & $\hlwhite{0.45} \vrule \hlwhite{0.47}$ & $\hlheavygreen{0.96} \vrule \hlheavygreen{0.98}$ & $\hlheavyred{0.03} \vrule \hlgreen{0.71}$ & $\hlheavygreen{0.96} \vrule \hlheavygreen{0.98}$ & $\hlheavygreen{0.97} \vrule \hlheavygreen{0.95}$ & $\hlheavygreen{0.94} \vrule \hlheavygreen{0.91}$ & $\hlwhite{0.59} \vrule \hlgreen{0.74}$ & $\hlheavygreen{0.95} \vrule \hlheavygreen{0.89}$ & $\hlheavygreen{0.94} \vrule \hlheavygreen{0.88}$ \\ \hline
 \Xhline{0.8pt}
    \end{tabular}
    }
    \caption{\llama's results of $\mathrm{IR}$ of experimental setting (index column) \vs controlled setting (parametric memory only) on LongMemEval-m (left number in each cell) and \lmb (right number in each cell).}
    \label{tab:improvement_rate_parametric_llama}
\end{table*}

\section{Comparative Experiments on Gemini and Llama}
\begin{figure*}[t]
    \centering
    \includegraphics[width=\linewidth]{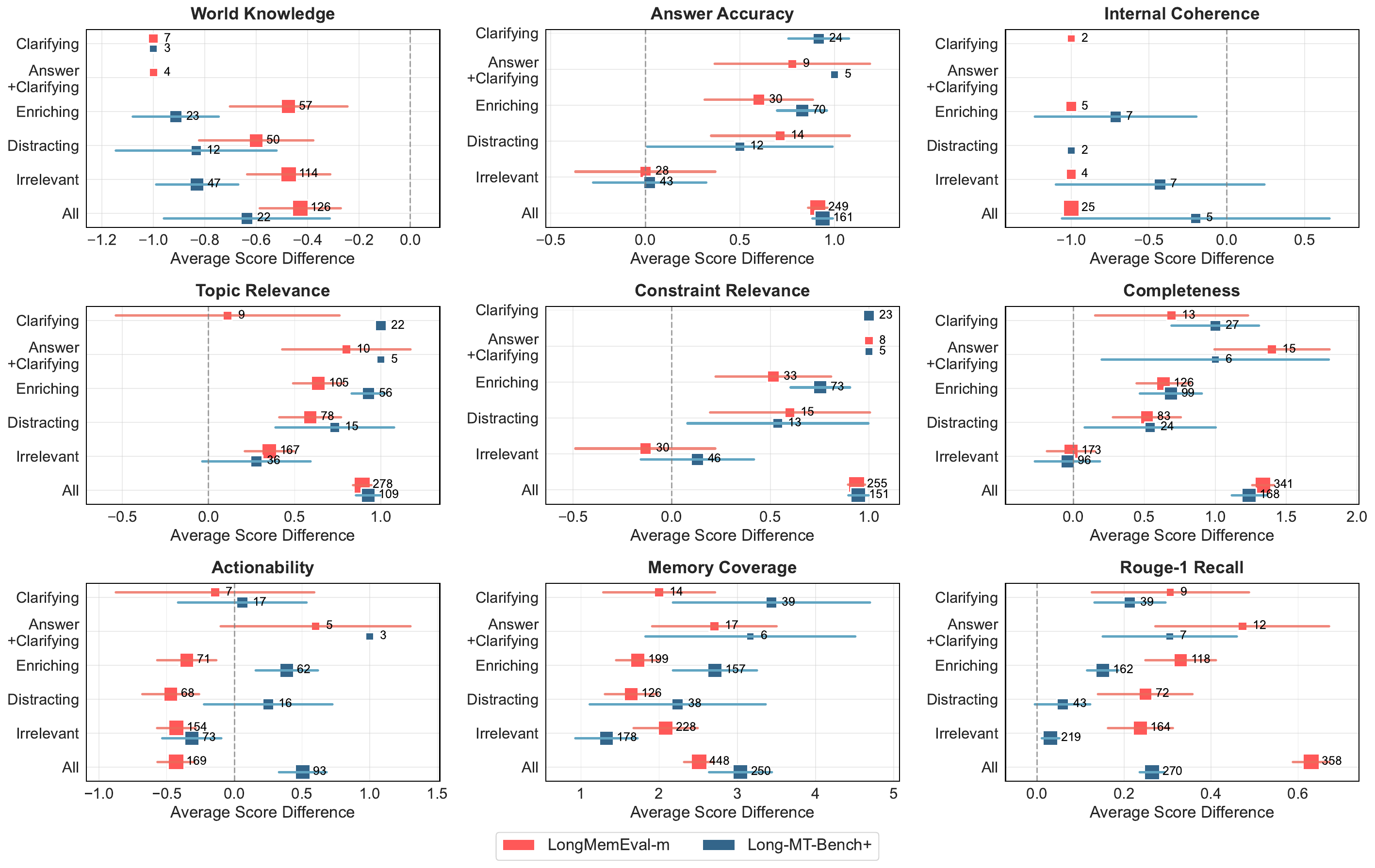}
    \caption{Results of mean per-query score differences with confident intervals when adding a memory setting relative to the \textbf{parametric-only baseline} on \gptshort. Squares mark the mean difference; horizontal bars show the confidence interval. Square size is proportional to the number of queries with a non-zero change, and the number shown beside each square equals that count.}
    \label{fig:blobbogram_parametric_gpt}
\end{figure*}

\begin{figure*}[t]
    \centering
    \begin{minipage}{\linewidth}
        \centering
        \includegraphics[width=\linewidth]{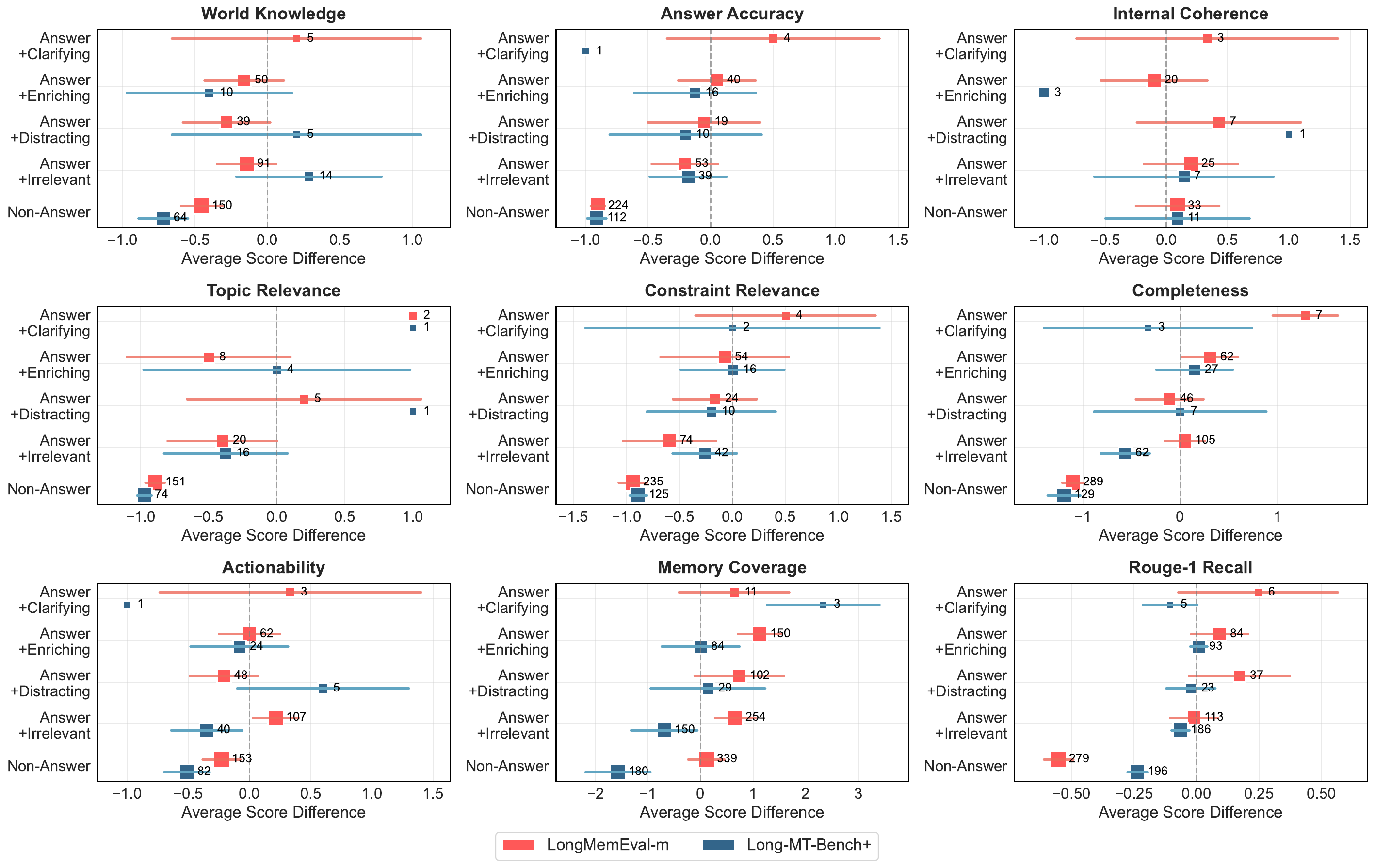}
        \vspace{-20pt}
        \caption{Results of mean per-query score differences with confident intervals when adding a memory setting relative to the \textbf{answer-only baseline} on \gemini. Squares mark the mean difference; horizontal bars show the confidence interval. Square size is proportional to the number of queries with a non-zero change, and the number shown beside each square equals that count.}
        \label{fig:blobbogram_answer_gemini}
    \end{minipage}
    \vfill
    \vspace{10pt}
    \begin{minipage}{\linewidth}
        \centering
        \includegraphics[width=\linewidth]{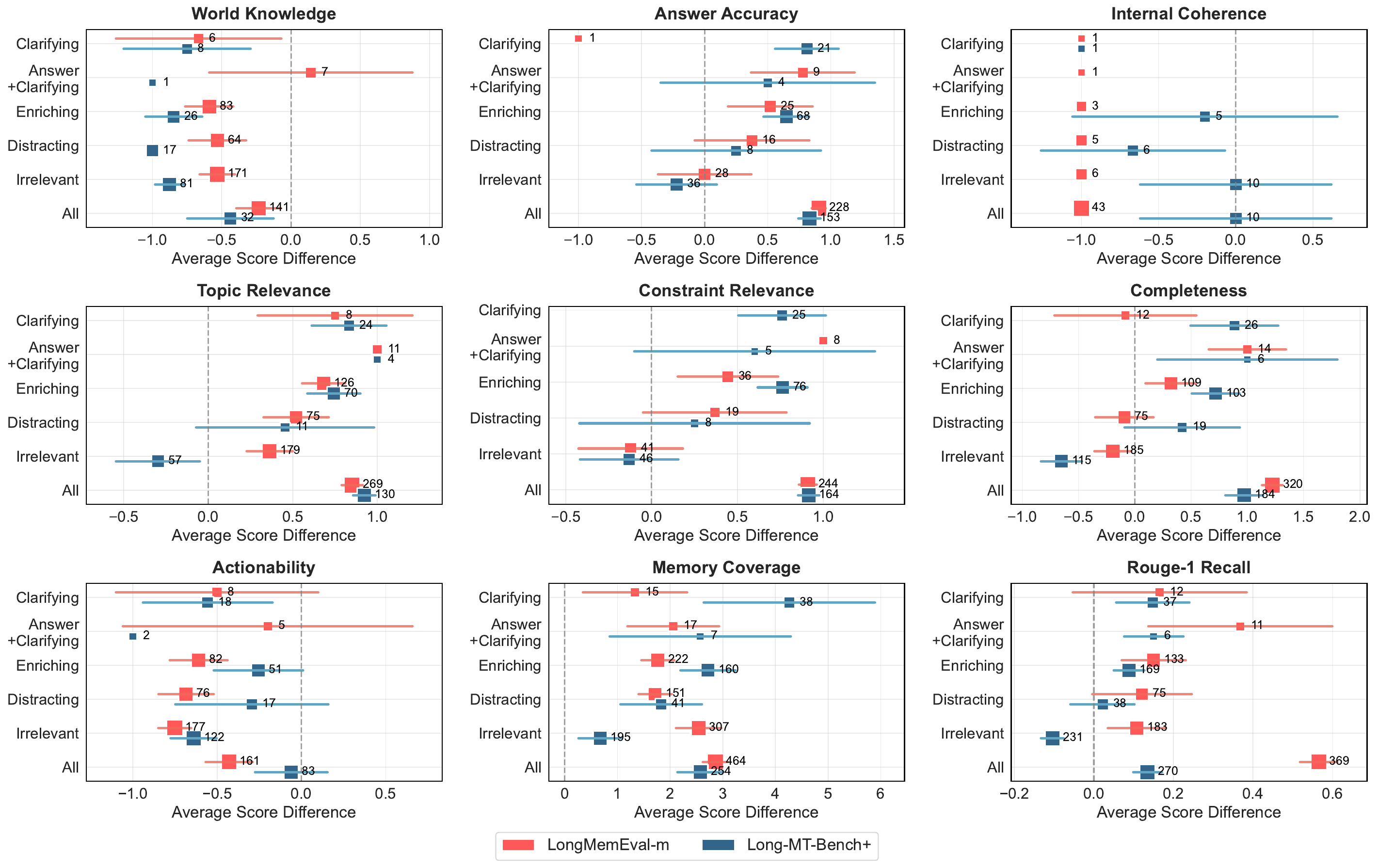}
        \vspace{-20pt}
        \caption{Results of mean per-query score differences with confident intervals when adding a memory setting relative to the \textbf{parametric-only baseline} on \gemini. Squares mark the mean difference; horizontal bars show the confidence interval. Square size is proportional to the number of queries with a non-zero change, and the number shown beside each square equals that count.}
        \label{fig:blobbogram_parametric_gemini}
    \end{minipage}
\end{figure*}

\begin{figure*}[t]
    \centering
    \begin{minipage}{\linewidth}
        \centering
        \includegraphics[width=\linewidth]{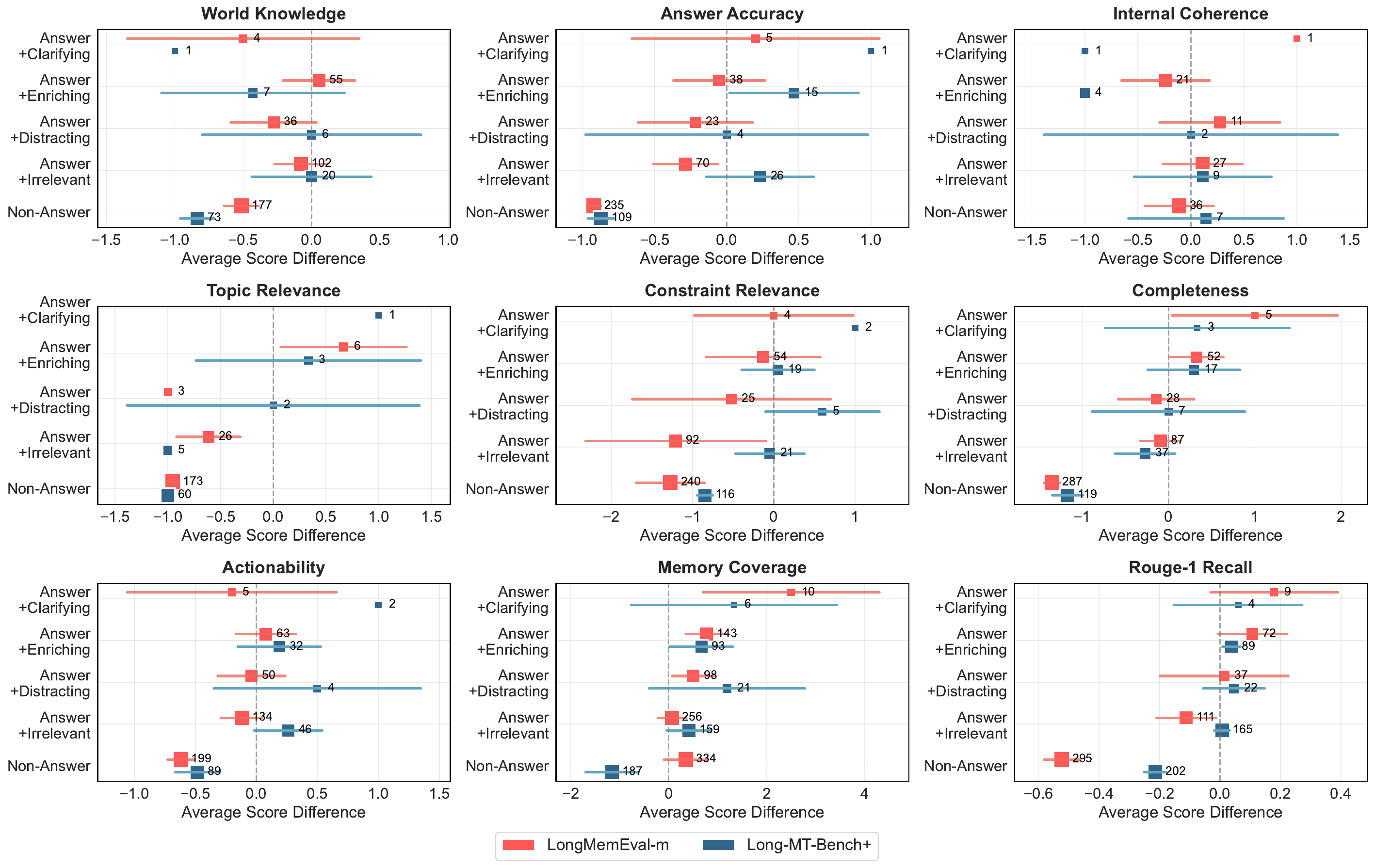}
        \caption{Results of mean per-query score differences with confident intervals when adding a memory setting relative to the \textbf{answer-only baseline} on \llama. Squares mark the mean difference; horizontal bars show the confidence interval. Square size is proportional to the number of queries with a non-zero change, and the number shown beside each square equals that count.}
        \label{fig:blobbogram_answer_llama}
    \end{minipage}
    \vfill
    \vspace{10pt}
    \begin{minipage}{\linewidth}
        \centering
        \includegraphics[width=\linewidth]{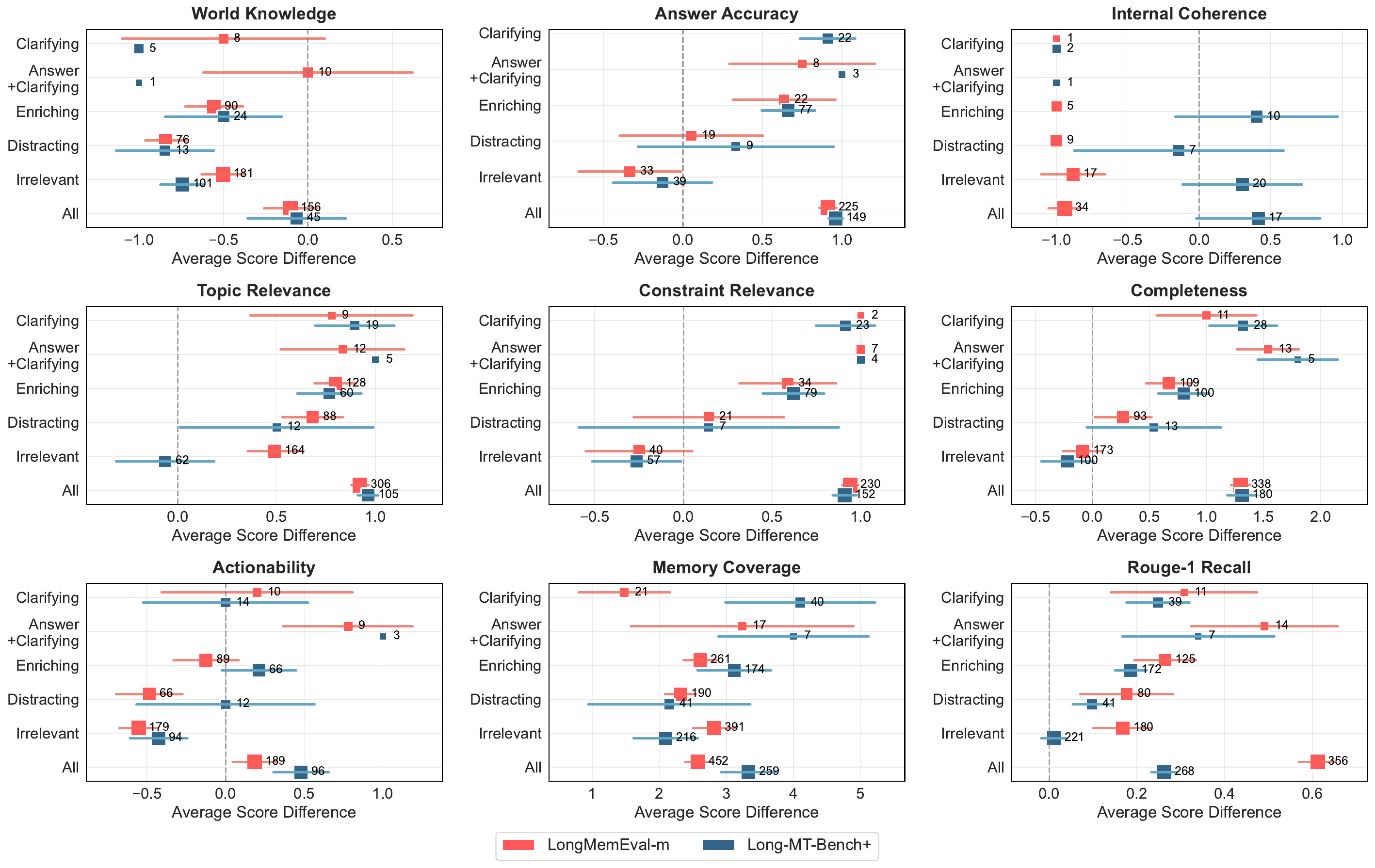}
        \caption{Results of mean per-query score differences with confident intervals when adding a memory setting relative to the \textbf{parametric-only baseline} on \llama. Squares mark the mean difference; horizontal bars show the confidence interval. Square size is proportional to the number of queries with a non-zero change, and the number shown beside each square equals that count.}
        \label{fig:blobbogram_parametric_llama}
    \end{minipage}
\end{figure*}

Table~\ref{tab:improvement_rate_answer_gemini}, ~\ref{tab:improvement_rate_answer_llama}, ~\ref{tab:improvement_rate_parametric_gemini} and~\ref{tab:improvement_rate_parametric_llama} show the improvement rate $\mathrm{IR}$ results of \gemini and \llama on different comparative experiments. Figure~\ref{fig:blobbogram_parametric_gpt} show the statistical results of $\mathrm{L}^m(i,j)$ of \gpt on experiment comparing with parametric memory only. Figure~\ref{fig:blobbogram_answer_gemini}, ~\ref{fig:blobbogram_parametric_gemini}, ~\ref{fig:blobbogram_answer_llama}, and~\ref{fig:blobbogram_parametric_llama} are statistical results of specific model and comparison setting.


\newpage
\begin{table*}[!ht]
  \centering
  \small
  \renewcommand{\arraystretch}{1.8}
    \setlength{\tabcolsep}{4pt}
    \resizebox{\textwidth}{!}{
    \begin{tabular}{| m{15cm} | m{1.5cm} |}
    \Xhline{0.8pt}
    \multicolumn{2}{|>{\raggedright\arraybackslash}p{16cm}|}{
    \textbf{Instruction}: ...
    
    \textbf{User query}: I'm planning to visit Bandung again and I was wondering if you could remind me of the name of that restaurant in Cihampelas Walk that serves a great Nasi Goreng? 

    \textbf{Gold answer}: Miss Bee Providore} \\ \Xhline{0.8pt}

    \textbf{Memory piece}:
    
    [User]: I think I'll start with Cihampelas Walk to check out the denim street. Can you recommend any good restaurants or cafes in the area? 
    
    [Assistant]: Sure! There are many good restaurants and cafes in the Cihampelas Walk area. Here are a few options: 1. Miss Bee Providore: This restaurant serves a mix of western and Indonesian cuisine and has a cozy and stylish interior. 2. Bakery \& Co: A great place to relax and enjoy a cup of coffee and a pastry. 3. Raa Cha Suki \& BBQ: If you're looking for something spicy and delicious, this Thai restaurant is a great option. 4. Sumoboo: This Japanese shaved ice dessert place features a wide variety of flavors, toppings, and combinations. 5. Holycow! Steakhouse by Chef Afit: This steakhouse has a warm atmosphere and serves high-quality steaks at reasonable prices. These are just a few of the many options you can explore in the area. Make sure to do some research and read reviews before making a final decision. Enjoy your visit! & Enriching \\ \hline
    \textbf{Memory piece}:

    [User]: I'm definitely going to try out Miss Bee Providore for some delicious food. Do they have any signature dishes I should try?
    
    [Assistant]: Miss Bee Providore offers a variety of delicious dishes, both Western and Indonesian cuisine. Here are a few signature dishes you should definitely try: 1. Miss Bee's Nasi Goreng: Their take on the classic Indonesian fried rice dish is a must-try! It's served with a side of prawn crackers and topped with a fried egg. 2. Yoghurt Granola Bowl: This is a great option if you're looking for something healthy and refreshing. It's made with homemade granola, fresh fruit, and yogurt. 3. Fried Chicken Burger: This signature burger features a juicy chicken patty, crispy bacon, lettuce, tomato, and a special aioli sauce. It's served with a side of fries. 4. Beef Rendang: If you're looking for a traditional Indonesian dish, try their version of beef rendang. It's a slow-cooked beef dish in a flavorful, spicy coconut milk sauce. 5. Chocolate Brownie: Finish off your meal with a sweet treat - Miss Bee's chocolate brownie is a local favorite and is served warm with vanilla ice cream and chocolate sauce. These are just a few of the many delicious options, so feel free to explore their menu and try something new! & Answer \\ \hline
    
    \textbf{Memory piece}:
    
    [User]: Yum, all of those dishes sound amazing! I'm definitely going to try the Miss Bee's Nasi Goreng and finish it off with the chocolate brownie.
    
    [Assistant]: Enjoy your meal at Miss Bee Providore and have a great time shopping at Cihampelas Walk! & Answer \\ \hline
    \textbf{Memory piece}:
    
    [User]: I'm so excited to explore more of Bandung! Do you have any recommendations for other fun activities to do in the city?
    
    [Assistant]: Sure, Bandung has plenty of fun activities to offer! Here are a few recommendations: 1. Visit a volcano: Bandung is located near several active and dormant volcanoes, including Tangkuban Perahu and Kawah Putih. You can take a day trip to visit these fascinating natural wonders. 2. Explore the tea plantations: Bandung is also known for its tea plantations, such as the Malabar Tea Plantation and the Puncak Tea Garden. You can take a tour of these plantations, enjoy a cup of tea, and enjoy the beautiful scenery. 3. Visit the Trans Studio Bandung: This is one of Indonesia's largest indoor theme parks, with rides, attractions, and entertainment for all ages. 4. Go shopping at the Paris Van Java mall: This mall is known for its luxurious and trendy boutiques, restaurants, and cafes, as well as its modern architecture. 5. Take a cooking class: Bandung is famous for its cuisine, and you can take a cooking class to learn how to make traditional Indonesian dishes such as nasi goreng, sate, and rendang. 6. Visit the Saung Angklung Udjo: This is a cultural center that features traditional music and dance performances with the angklung, a traditional Indonesian instrument made from bamboo. These are just a few of the many activities that you can do in Bandung. Make sure to do some research and plan your trip accordingly, depending on your interests and preferences. Have fun exploring! & Irrelevant \\ \hline
    \textbf{Memory piece}:
    
    [User]: What are some unique shopping experiences to be had in Bandung? 
    
    [Assistant]: As an AI language model, I don't have personal experiences, but based on research, here are some unique shopping experiences to be had in Bandung: 1. Factory outlets: Bandung is well known for being a factory outlet (FO) paradise. You can find many factory outlets selling local and international brands at discounted prices. 2. Pasar Baru: This is Bandung's oldest and largest traditional market, selling everything from clothing and accessories to food and home goods. It's a great place to experience the hustle and bustle of a traditional market. 3. Braga Street: This historic street is lined with old colonial buildings that have been turned into trendy cafes, restaurants, and boutique shops. It's a great place to find unique souvenirs and gifts. 4. Cihampelas Walk: Also known as Ciwalk, this shopping center is famous for its denim street, where you can find a wide range of denim products, from jeans to jackets. 5. Dago Street: This street is home to many boutiques, art galleries, and cafes, making it a popular spot for young and trendy shoppers. 6. Rumah Mode: This popular factory outlet is famous for its Balinese-inspired architecture and stylish clothing and accessories. 7. Kartika Sari: This bakery is famous for its chocolate brownies and other Indonesian snacks, making it a great place to pick up some tasty treats to take home. & Irrelevant \\
    \Xhline{0.8pt}
    \end{tabular}
    }
  \caption{A judgment case for \texttt{GPT-5-reasoning} to determine the memory type. Left column is each memory piece and right column is the determined type. See the full prompt in Figure~\ref{fig:prompt_memory_type}.}
  \label{tab:judgment_case_1}
\end{table*}
\begin{table*}[!ht]
  \centering
  \small
  \vspace{-5pt}
  \renewcommand{\arraystretch}{1.8}
    \setlength{\tabcolsep}{4pt}
    \resizebox{\textwidth}{!}{
    \begin{tabular}{| m{15.2cm} | m{1.5cm} |}
    \Xhline{0.8pt}
    \multicolumn{2}{|>{\raggedright\arraybackslash}p{16cm}|}{
    \textbf{Instruction}: ...
    
    \textbf{User query}: I'm planning to revisit Orlando. I was wondering if you could remind me of that unique dessert shop with the giant milkshakes we talked about last time? 

    \textbf{Gold answer}: The Sugar Factory at Icon Park.} \\ \Xhline{0.8pt}

    \textbf{Memory piece}:
    
    [User]: Wow, these dining options sound amazing! Do you have any recommendations for a fun dessert spot that my family can check out after dinner?
    
    [Assistant]: Absolutely! Here are some fun dessert spots that your family might enjoy after dinner: 1. The Sugar Factory - A sweet shop located at Icon Park that offers an enormous menu of sweet treats, including specialty drinks and giant milkshakes. 2. Wondermade - A gourmet marshmallow shop located in Sanford, just north of Orlando, that offers unique flavored marshmallows and s'mores kits. 3. Gideon's Bakehouse - A bakery located at Disney Springs that offers delicious cookies, cakes, and other baked treats. 4. Kelly's Homemade Ice Cream - A small-batch ice cream shop located in Orlando that offers unique flavors like Banana Cream Pie and Strawberry Jalapeno. 5. Voodoo Doughnut - A doughnut shop located at Universal CityWalk that offers imaginative and creative doughnuts with unique toppings and flavors. 6. The Glass Knife - A dessert and coffee shop located in Winter Park that serves up beautiful cakes, pastries, and other sweet treats in a stylish setting. 7. Lickety Splitz - A dessert shop located in Orlando that serves up handmade frozen yogurt, sorbet, and ice cream. 8. Better Than Sex - A dessert shop located in the Ivanhoe Village neighborhood of Orlando that specializes in decadent desserts like chocolate fondue and cheesecake. 9. The Greenery Creamery - An ice cream shop located in the Thornton Park neighborhood of Orlando that offers all-natural ice cream flavors like lavender, honeycomb, and matcha. 10. The Pastry Studio - A bakery located in Daytona Beach, just east of Orlando, that offers beautiful and delicious custom cakes and other desserts. & Answer \\ \hline
    
    \textbf{Memory piece}:
    
    [User]: Can you recommend any places to eat in Orlando that are family-friendly?
    
    [Assistant]: Certainly! Orlando has many family-friendly dining options. Here are some recommendations: 1. Toothsome Chocolate Emporium \& Savory Feast Kitchen - A restaurant located at Universal CityWalk that offers a unique dining experience with imaginative dishes and delicious desserts. 2. The Cheesecake Factory - A casual dining restaurant located in The Mall at Millenia that offers a diverse menu that includes everything from burgers to salads and, of course, an extensive selection of cheesecakes. 3. Shake Shack - A fast-food restaurant located in Winter Park that offers burgers, hot dogs, and shakes. 4. Hash House A Go Go - A restaurant located on International Drive that serves up \"twisted farm food\" that is both delicious and visually stunning. 5. Bahama Breeze - A Caribbean-themed restaurant located on International Drive that has a family-friendly atmosphere and offers a menu of tropical dishes and drinks. 6. Blaze Pizza - A fast-casual restaurant located in Disney Springs that offers build-your-own Neapolitan pizzas and salads. 7. T-Rex Cafe - A dinosaur-themed restaurant located in Disney Springs that offers a unique prehistoric dining experience for families. 8. Uncle Julio's - A Tex-Mex restaurant located in Orlando and Kissimmee that offers a menu of authentic Mexican dishes and delicious drinks. 9. Pio Pio - A restaurant located in the Dr. Phillips neighborhood that serves up delicious Peruvian dishes. 10. Sweet Tomatoes - A buffet restaurant chain located throughout Orlando that offers a variety of healthy and delicious salads, soups, and baked goods. & Distracting \\ \hline

    \textbf{Memory piece}:

    [User]: Wow, those tips are great! I'm excited to take on this dessert adventure with my family. Do you have any more recommendations for other fun and unique things to do in Orlando?
    
    [Assistant]: Yes, there are plenty of other fun and unique things to do in Orlando! Here are some more recommendations: 1. Escape Rooms - Orlando has several escape rooms that offer immersive and challenging puzzle-solving experiences. 2. Cirque du Soleil - Watch a mesmerizing acrobatic and theatrical performance by the world-famous Cirque du Soleil, which often performs in Orlando. 3. Airboat Tours - Take an airboat tour through the Florida Everglades and see alligators, birds, and other wildlife up close. 4. The Orlando Eye - Take a ride on the 400-foot observation wheel and enjoy panoramic views of Orlando. 5. Orlando Science Center - Visit the Orlando Science Center, which has interactive exhibits and educational programs that entertain visitors of all ages. 6. The Orlando Museum of Art - See beautiful artwork from various periods and artists at the Orlando Museum of Art. 7. Amusement Parks - Orlando has many amusement parks besides Disney and Universal, such as Fun Spot America, ICON Park, and SeaWorld. 8. The Charles Hosmer Morse Museum of American Art - See beautiful works of art by Louis Comfort Tiffany, including iconic stained glass lamps. 9. Dr. Phillips Center for the Performing Arts - Enjoy a variety of dance, music and theater performances at this beautiful venue. 10. The Orlando Shakespeare Theater - Experience the theater with classic and contemporary works in a unique setting. & Irrelevant \\ \hline
    
    \textbf{Memory piece}:
    
    [User]: I don't really trust buffet restaurants. Can you suggest any other family-friendly dining options in Orlando?
    
    [Assistant]: Sure, here are some additional family-friendly dining options in Orlando: 1. The Cowfish Sushi Burger Bar - A unique restaurant located at Universal CityWalk that specializes in creative sushi and gourmet burgers. 2. BJ's Restaurant and Brewhouse - A casual dining restaurant located in Orlando that offers a diverse menu of American style dishes and beers brewed on site. 3. Mediterranean Blue - A Greek restaurant located in Orlando that serves up classic Greek dishes like souvlaki and gyros. 4. Black Bean Deli - A casual Cuban restaurant located in Orlando that serves up delicious Cuban sandwiches, plantains, and more. 5. Toothsome Chocolate Emporium \& Savory Feast Kitchen - As mentioned before, this is a great family-friendly option for lunch or dinner. 6. Dandelion Communitea Cafe - A vegetarian and vegan restaurant located in Orlando that serves up fresh and healthy salads, sandwiches, and smoothies. 7. The Whiskey - A restaurant located in Orlando that has a family-friendly outdoor seating area and offers a menu of American dishes with a Southern twist. 8. Blaze Pizza - Another option mentioned before, this fast-casual pizzeria is great for a quick meal. 9. Kobe Japanese Steakhouse - A Japanese hibachi grill restaurant located in Orlando that offers an entertaining dining experience and delicious seafood, meat, and vegetable dishes. 10. Soco Restaurant - A restaurant located in the Thornton Park neighborhood of Orlando that offers Southern-style dishes with a modern twist. They also have a great weekend brunch. & Distracting \\ 
    \Xhline{0.8pt}
    \end{tabular}
    }
    \caption{A judgment case for \texttt{GPT-5-reasoning} to determine the memory type. Left column is each memory piece and right column is the determined type. See the full prompt in Figure~\ref{fig:prompt_memory_type}.}
  \label{tab:judgment_case_2}
\end{table*}
\begin{table*}[!ht]
  \centering
  \small
  \vspace{-5pt}
  \renewcommand{\arraystretch}{1.8}
    \setlength{\tabcolsep}{4pt}
    \resizebox{\textwidth}{!}{
    \begin{tabular}{|c|c|c|c|c|c|c|}
    \Xhline{0.8pt}
    \multicolumn{7}{|p{16cm}|}{
    \textbf{Instruction}: ...

    \textbf{Prior conversation history}: ...
    
    \textbf{User query}: I wanted to check back on our previous conversation about Netflix. I mentioned that I wanted to be able to access all seasons of old shows? Do you remember what show I used as an example, the one that only had the last season available?

    \textbf{Gold answer}: Doc Martin.} \\ \Xhline{0.8pt}

    \multicolumn{7}{|p{16cm}|}{\textbf{Answer} ($k=0$): I'm sorry, I don't have the ability to recall past conversations. However, if you let me know the name of the show again, I can try to help you with any information or updates regarding its availability on Netflix.} \\ \hline
    World Knowledge & Answer Accuracy & Internal Coherence & Topic Relevance & Constraint Awareness & Completeness & Actionability \\ \hline
    1 & 0 & 1 & 1 & 0 & 0 & 0 \\ \Xhline{0.8pt}
    \multicolumn{7}{|p{16cm}|}{\textbf{Answer} ($k=5$):
    Yes, I remember! You mentioned the show **"Doc Martin"** as an example where only the last season was available on Netflix. If you have any more questions or need further assistance, feel free to ask!} \\ \hline
    1 & 1 & 1 & 1 & 1 & 2 & 0 \\ \Xhline{0.8pt}
    \multicolumn{7}{|p{16cm}|}{\textbf{Answer} ($k=10$):
    Yes, I remember! You mentioned that you wanted to have access to all seasons of old shows on Netflix, and you used **"Doc Martin"** as an example, noting that only the last season was available.} \\ \hline
    1 & 1 & 1 & 1 & 1 & 2 & 0 \\ \Xhline{0.8pt}
    \multicolumn{7}{|p{16cm}|}{\textbf{Answer} ($k=20$):
    Yes, I remember! You mentioned that you wanted to have access to all seasons of old shows and used **"Doc Martin"** as an example, noting that only the last season was available.} \\ \hline
    1 & 1 & 1 & 1 & 1 & 2 & 0 \\ \Xhline{0.8pt}
    \multicolumn{7}{|p{16cm}|}{\textbf{Answer} ($k=50$):
    Yes, you mentioned that you wanted to be able to access all seasons of old shows on Netflix, and you used **"Doc Martin"** as an example, noting that it only had the last season available. If you have any more questions or need further assistance, feel free to ask!} \\ \hline
    1 & 1 & 1 & 1 & 1 & 2 & 0 \\ \Xhline{0.8pt}
    \multicolumn{7}{|p{16cm}|}{\textbf{Answer} ($k=100$):
    Yes, I remember our conversation about Netflix. You mentioned that you wanted to be able to access all seasons of old shows and used **"Doc Martin"** as an example, noting that it went down after a while, and now only the last season is available. If you have any more questions or need further assistance, feel free to ask!} \\ \hline
    1 & 1 & 1 & 1 & 1 & 2 & 0 \\ 
    \Xhline{0.8pt}
    \end{tabular}
    }
  \caption{A judgment case for \texttt{GPT-3o} to score the response on different metrics. We provide responses under different retrieval size ($k$) and their scores.}
  \label{tab:judgment_case_3}
\end{table*}
\label{sec:appendix}

\end{document}